\documentclass[lettersize,journal]{IEEEtran}
\usepackage{amsmath,amsfonts}
\usepackage{algorithmic}
\usepackage{algorithm}
\usepackage{array}
\usepackage[caption=false,font=normalsize,labelfont=sf,textfont=sf]{subfig}
\usepackage{textcomp}
\usepackage{stfloats}
\usepackage{url}
\usepackage{verbatim}
\usepackage{graphicx}
\usepackage{cite}
\usepackage{caption}
\usepackage{booktabs} 

\usepackage{graphicx}  
\usepackage{subfig}    
\usepackage{float}     
\usepackage{caption}
\usepackage{graphicx}
\usepackage{float} 
\usepackage{subcaption}
\usepackage{hyperref}

\begin{document}

\title{YOLO-Ant: A Lightweight Detector via Depthwise Separable Convolutional and Large Kernel Design for Antenna Interference Source Detection}

\author{		
	Xiaoyu Tang,~\IEEEmembership{Member,~IEEE,}
	Xingming Chen,
	Jintao Cheng,
	
	Jin Wu,~\IEEEmembership{Member,~IEEE,}
	Rui Fan,~\IEEEmembership{Senior Member,~IEEE,}
	Chengxi Zhang,~\IEEEmembership{Member,~IEEE,} Zebo Zhou

\thanks{This research was supported by the National Natural Science Foundation of China under Grant 62001173, the Project of Special Funds for the Cultivation of Guangdong College Students' Scientific and Technological Innovation (“Climbing Program” Special Funds) under Grant pdjh2022a0131 and pdjh2023b0141, the National Natural Science Foundation of China under Grant 42074038, the Fundamental Research Funds for the Central Universities and Xiaomi Young Talents Program.}
        
\thanks{Corresponding author: Xiaoyu Tang. E-mail address: tangxy@scnu.edu.cn.}
\thanks{Xiaoyu Tang, Xingming Chen and Jintao Cheng is with the School of Electronic and Information Engineering, Faculty of Engineering, South China Normal University, Foshan, Guangdong 528225, China, and also with the School of Physics, South China Normal University, Guangzhou, Guangdong 510000, China.
	
Jin Wu is with the Department of Electronic and Computer Engineering, Hong Kong University of Science and Technology, Hong Kong.

Rui Fan is with the College of Electronics \& Information Engineering, Shanghai Research Institute for Intelligent Autonomous Systems, the State Key Laboratory of Intelligent Autonomous Systems, and Frontiers Science Center for Intelligent Autonomous Systems, Tongji University, Shanghai 201804, China (e-mail: rui.fan@ieee.org).

Chengxi Zhang is with the School of Internet of Things Engineering, Jiangnan University, Wuxi, 214122, China.

Zebo Zhou is with the School of Aeronautics \& Astronautics University of Electronic Science and Technology of China and Aircraft swarm intelligent sensing and cooperative control Key Laboratory of Sichuan Province, Chengdu 610097, China, and also with the National Laboratory on Adaptive Optics, Chengdu 610209, China.
}}

\markboth{IEEE TRANSACTIONS ON INSTRUMENTATION AND MEASUREMENT,~Vol.~14, No.~8, August~2021}%
{Shell \MakeLowercase{\textit{et al.}}: A Sample Article Using IEEEtran.cls for IEEE Journals}

\IEEEpubid{0000--0000/00\$00.00~\copyright~2021 IEEE}

\maketitle

\begin{abstract}
In the era of 5G communication, removing interference sources that affect communication is a resource-intensive task. The rapid development of computer vision has enabled unmanned aerial vehicles to perform various high-altitude detection tasks. Because the field of object detection for antenna interference sources has not been fully explored, this industry lacks dedicated learning samples and detection models for this specific task. In this article, an antenna dataset is created to address important antenna interference source detection issues and serves as the basis for subsequent research. We introduce YOLO-Ant, a lightweight  CNN and transformer hybrid detector specifically designed for antenna interference source detection. Specifically, we initially formulated a lightweight design for the network depth and width, ensuring that subsequent investigations were conducted within a lightweight framework. Then, we propose a DSLK-Block module based on depthwise separable convolution and large convolution kernels to enhance the network’s feature extraction ability, effectively improving small object detection. To address challenges such as complex backgrounds and large interclass differences in antenna detection, we construct DSLKVit-Block, a powerful feature extraction module that combines DSLK-Block and transformer structures. Considering both its lightweight design and accuracy, our method not only achieves optimal performance on the antenna dataset but also yields competitive results on public datasets.
\end{abstract}


\section{Introduction}
\begin{figure}[!t]
	\centering
	\includegraphics[scale=0.8]{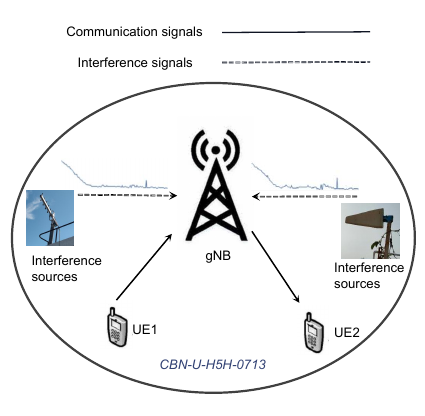}
	\caption{The process of 5G communication in the CBN-U-H5H-0713 area is shown in the figure. Two antenna interference source signals appear in it. The gNB (gNodeB) denotes a 5G base station. The UE (User Equipment) denotes the terminal equipment that users use to access the wireless network.}
 \vspace{-14pt} 
\label{figure1}
\end{figure}

\IEEEPARstart{T}{o} ensure high-quality communication in people's work and daily lives, various wireless devices operate in different frequency bands. 5G communication is of particular note due to its introduction of new frequency bands into everyday communication. However, due to the presence of numerous private wireless signals that have not undergone spectrum allocation by communication regulatory authorities, the 5G communication network has accumulated a considerable number of sources of interference. If individuals operate in the same geographical areas and occupy similar or adjacent frequency bands as these interference signals in their everyday communication, this will result in a significant deterioration in communication quality, as shown in Fig. \ref{figure1}. Regular remediation of radio interference sources is vital for communication departments to alleviate this situation. The identification of signal interference sources necessitates monitoring personnel to visually inspect areas where communication quality is compromised due to the presence of suspicious antennas elevated at high altitudes, constituting a time-consuming and labor-intensive task.
In light of the mature advancements in unmanned aerial vehicle (UAV) cruising technology and object detection techniques within computer vision, unmanned drones have become viable alternatives for handling complex and challenging detection tasks previously performed by humans. For example,\cite{lu2021attention} \cite{yang2022bidirection} \cite{hao2022insulator} noted that object detection tasks in deep learning combined with UAVs have been useful in production and other areas. The success of these approaches has demonstrated the feasibility of utilizing UAVs for object detection tasks aimed at interference source antennas. However, due to the nascent stage of this detection task within the current domain of object detection, the creation of a suitable antenna dataset and the exploration of appropriate object detection methodologies are of paramount importance. 
\IEEEpubidadjcol

Convolutional architectures are the basis for most object detection frameworks in industrial scenarios and rely on the development of efficient convolutional neural networks in deep learning. When addressing various tasks and technical challenges, corresponding enhancements to these architectures are necessary. The antenna interference source object detection task presents three main challenges. The first issue pertains to the lightweight nature and low computational complexity of a detection model; consequently, object detectors can be deployed on lightweight computing devices, enabling real-time object detection via UAVs. 
Previous research, exemplified by GhostNet\cite{han2020ghostnet} and EfficientNet\cite{tan2021efficientnetv2}, has focused on designing lightweight networks as potential backbones for different detection models to achieve an overall lightweight solution. However, these networks are susceptible to information feature loss. The second difficulty in the antenna interference source object detection task lies in the differences arising from the different angles and heights from which the UAV captures the antennas. These variations result in a nonuniform distribution of target sizes within the images, most of which are extremely small in size. Additionally, there is a significant interclass dissimilarity issue, wherein antennas of the same type exhibit markedly different morphologies in different images. To address these issues,  researchers have explored two aspects: multiscale feature learning and attention mechanisms \cite{lim2021small}\cite{yang2019scrdet}\cite{yang2022querydet}\cite{JI2023108490}.  However, despite improvements in small object detection accuracy, these methods encounter challenges related to the model's weak generalizability and robustness. As a result, the overall detection accuracy for the targets was compromised. Moreover, the computational complexity of such models is higher. The third difficulty is complex target backgrounds, which cause serious false and missed detections. Given that antennas are commonly installed on tall buildings or fenced balconies in practical scenarios, the resulting complex and mutually obscuring environment between the target and the background significantly hinders detection. Researchers have suggested using attention or self-attention mechanisms to address this difficulty. In \cite{2019GCNet}\cite{carion2020end}\cite{zheng2021rethinking}, a squeeze-and-excitation(SE) attention module was proposed, or a self-attention structure was used to build the whole network for object detection. The advantage of these models lies in their ability to effectively capture the spatial relationship between the target and the background. This capability significantly enhances object detection performance on complex backgrounds. However, these mechanisms tend to consume considerable computational resources and memory, which is not consistent with the original lightweight design intention. Additionally, networks built solely on self-attention mechanisms also suffer from long training times and poor detection accuracy for small objects.

In response to the aforementioned limitations, we propose YOLO-Ant, a lightweight one-stage detector designed for detecting antenna interference sources with small targets and complex backgrounds. 
Initially, we analyze the scale and number of channels in each feature layer of the model; subsequently, we design the network's width and depth to ensure that the entire detection process is performed within a lightweight framework. Our design considerations aim to balance detection accuracy with the reduction of model parameters and computational complexity.
To address the issues of small target size and large interclass variation, we implement an efficient feature extraction module based on depthwise separable convolution, DSLK-Block, which is applied to each feature layer in the model. 
This method effectively enhances the network's feature learning and fusion capabilities, leading to a significant improvement in detection accuracy for all types of targets, particularly small targets. Additionally, this approach contributes to reducing the model's overall weight. Finally, to address the problem of complex backgrounds, YOLO-Ant uses an innovative CNN and transformer hybrid structure to act on the neck of the model. This process enables us to fully utilize both local and global feature learning to address the challenges posed by complex backgrounds while still accounting for small object detection. This approach significantly improves all the detection accuracy indicators while only slightly increasing the model’s number of parameters and computational complexity. To demonstrate the model’s generalizability and robustness, we also tested YOLO-Ant on public datasets and achieved highly competitive results. In conclusion, the main contributions of this paper are as follows:
\begin{enumerate}{}{}
	\item{In response to the lack of learning samples for antenna object detection schemes, we conducted image acquisition and manual annotation of the three most common types of antennas encountered in real-world interference source investigation tasks. This dataset is pioneering and establishes the foundation for subsequent work.}
	\item{We initially pruned YOLOv5-s \cite{glenn_jocher_2021_5563715}, obtaining a lightweight detection framework. Within this framework, (i) a lightweight plug-and-play module based on depthwise separable convolution combined with large convolutional kernels was proposed to effectively improve the feature extraction and detection capabilities of the network for small targets; (ii) the innovative use of a transformer module to construct the neck structure of the detection model improved the detection capability of the network without increasing the model’s parameter count or computational complexity, effectively solving the problem of dealing with complex backgrounds.}
	\item{Our proposed method achieves state-of-the-art (SOTA) performance on the antenna dataset, striking a balance between lightweight design and detection accuracy. Moreover, YOLO-Ant yields competitive results on public datasets such as COCO, validating its robustness and superior performance.  Source code is released in \href{ https://github.com/SCNU-RISLAB/YOLO-Ant}{https://github.com/SCNU-RISLAB/YOLO-Ant}.}
\end{enumerate}

The  remainder of this paper is structured as follows. Section II presents related work, briefly introducing the improvement points of the model proposed in this paper and the related work involved, including CNN network development, the emergence of the transformer detection model and the crossover development between the CNN and transformer. Section III discusses the proposed lightweight detection framework based on the YOLOv5-s improvement, including pruning the baseline model, the design of the DSLK-Block module, and the neck structure built based on DSLKVit-Block. The experimental results are given in Section IV, and Section V concludes the paper.

\section{RELATED WORK}
\subsection{CNN (convolutional neural network)-based object detection}
The development of object detection in the computer vision field has been greatly influenced by CNN-based methods. Traditional approaches using hand-designed features and classifiers have been shown to be  inadequate, leading to the dominance of CNN-based methods. The initial CNN model, LeNet-5\cite{lecun1998gradient}, was limited by computational resources and model size. However, with advancements in computational power and larger datasets, deeper and more complex CNN models, such as AlexNet\cite{krizhevsky2017imagenet}, VGGNet\cite{simonyan2014very}, GoogLeNet\cite{szegedy2015going}, and ResNet\cite{he2016deep},  have emerged. These models have improved network accuracy, reduced parameters, and addressed network degradation issues, laying a solid foundation for 2D object detection.

Two distinct methods have emerged from the convoluted development of 2D object detection: two-stage and one-stage detectors. Two-stage detectors, such as R-CNN\cite{girshick2014rich} and Fast R-CNN\cite{girshick2015fast}, generate candidate frames using algorithms and perform classification and regression on each candidate frame. Faster R-CNN\cite{faster2015towards} introduces the region proposal network (RPN) for candidate frame generation. In contrast, one-stage detectors, such as YOLO\cite{redmon2016you} and SSD\cite{liu2016ssd}, perform classification and regression directly on each location in the input image. YOLOv2\cite{redmon2017yolo9000} and YOLOv3\cite{redmon2018yolov3} improved detection accuracy through methods such as multiscale prediction, batch normalization, and feature pyramid networks (FPNs). SSD introduces multiscale detection using multiple-scale feature maps, while RetinaNet\cite{lin2017focal} focuses on addressing the category imbalance problem.
For the aforementioned model, one-stage detectors are more suitable for real-time detection tasks on UAVs than two-stage detectors are because they do not require additional networks or algorithms for fine-tuning. However, to compensate for the deficiency in accuracy resulting from the pursuit of detection speed, improvements need to be made to the backbone and neck of the one-stage detector by developing various efficient feature extraction modules or structures.
The backbone and neck are the basic components of object detection models. The backbone is a CNN trained on image classification datasets such as ImageNet\cite{russakovsky2015imagenet}, in which the input image is transformed into a high-dimensional feature representation. The neck module further processes the feature map, changing the scale and resolution to extract different levels of feature information. Numerous object detection models, such as  NAS-FPN\cite{ghaisilearning}, EfficientDet\cite{tan2020efficientdet}, YOLOv4\cite{bochkovskiy2020yolov4}, and YOLOv7\cite{wang2022yolov7}, have been developed based on these concepts, incorporating various improvements and techniques to enhance accuracy and performance. However, these general models are often designed with modules that consider various common tasks, exhibiting generalizability but not effectively addressing specific challenges in particular scenarios. For instance, there are several challenges, such as small object detection and complex backgrounds, in our task. Therefore, making task-specific modifications is crucial when contemplating different tasks.

\subsection{Developing an attention mechanism in the CV domain}
Attention mechanisms, initially utilized in natural language processing, have gained significant traction in computer vision \cite{li2023roadformer,fan2023autonomous}, particularly in the field of object detection. Attention mechanisms such as channel attention, spatial attention, and their combinations have been introduced\cite{hu2018squeeze} \cite{woo2018cbam} \cite{fu2019dual}. They effectively utilize global and local information in feature maps, improving feature representation and attention weighting, thereby enhancing model accuracy and efficiency. However, for these conventional attention mechanisms, a fixed window size or other constraints are typically employed to regulate the correlation between each position and others. In contrast, self-attention mechanisms can extract information from different positions in the information sequence more flexibly, enabling the extraction of global information. This flexibility has contributed to the widespread application of transformer\cite{vaswani2017attention} models based on self-attention mechanisms, including in various domains such as computer vision. For example, the Vision Transformer (ViT)\cite{dosovitskiy2020image} splits images into patches for self-attention computations. The swin transformer\cite{liu2021swin} improves local information processing by using a window-based partitioning approach. Detection with transformers (DETR)\cite{carion2020end} adopts a global self-attention mechanism, allowing each position to obtain contextual information from the entire image. 
Naturally, transformers incur substantial computational costs and training time, posing challenges for model convergence. To address these challenges, researchers have introduced lightweight transformer object detectors, including MobileViT \cite{mehta2021mobilevit} and EdgeViT\cite{pan2022edgevits}. Moreover, innovative approaches such as conditional DETR\cite{meng2021conditional} and DN-DETR\cite{li2022dn}  have been developed to address the crucial issue of slow training convergence. However, due to their simplified design, the majority of current lightweight transformer structures are applicable only to classification tasks involving small-sized image inputs and are not suitable for detection tasks. The proposed detection methods aimed at addressing slow convergence have made transformer models more complex. Therefore, achieving a balance between the lightweight nature of transformer models and detection accuracy remains a crucial research scope in the current field of computer vision.

\subsection{Combination of CNN and Transformer}
In object detection, CNNs and transformers have distinct applications and advantages. CNNs are known for their strong image feature extraction abilities, ability to perform multichannel processing, and ability to learn spatial correlations. However, CNN-based models have limitations in handling objects of different sizes and proportions due to fixed window sizes and strides. On the other hand, transformers exhibit excellent performance in capturing long-range dependencies within input sequences without prior knowledge, albeit at a slower speed and requiring substantial amounts of training data. Evidently, the amalgamation of CNNs and transformers offers complementarity across various dimensions, and researchers have already delved into numerous methodologies to explore this synergy.

The pioneering DETR model replaces fully connected and convolutional layers with transformers while using ResNet as the feature extractor, improving accuracy and efficiency. Huawei's CMTBlock combines depthwise separable convolution and the transformer's multihead self-attention module for local and global information fusion. The CMT model\cite{guo2022cmt} stacks the CMTBlock in a hybrid CNN-transformer structure. The Conformer\cite{peng2021local} adopts a dual-network structure, where the CNN branch enhances local perception of the transformer branch. The mobile-former\cite{chen2022mobile} features parallel CNN and transformer modules with bidirectional bridges, leveraging MobileNet\cite{howard2017mobilenets} for local processing and the transformer for global interaction.
However, networks or models employing such hybrid structures face challenges in effectively balancing accuracy and lightweight design. For instance, detectors such as DETR, lacking FPN structures, exhibit suboptimal performance in small object detection. While the CMT and Conformer networks have proven effective in classification tasks, their application to downstream tasks such as object detection deviates from the realm of lightweight design. In contrast to the aforementioned models, which concatenate both structures, an alternative approach involves making transformer-style improvements directly on the CNN network. ConvNeXt\cite{liu2022convnet} implements novel architectures and optimization strategies similar to those of transformers, achieving competitive results without attention structures.  RepLKNet\cite{ding2022scaling} employs large convolutional kernels to widen the receptive field, thus emulating the transformer-like capability for global feature extraction. By investigating the computational principles of transformers, ACMix\cite{pan2022integration} maps their operation process onto convolutional operators, thereby combining them with traditional convolution operations to construct a novel CNN architecture. Parc-Net\cite{zhang2022parc} introduces circular convolution for global information extraction within a pure convolutional structure. Although these innovative networks may not achieve SOTA performance, their greater significance lies in exploring the factors contributing to the success of transformers from a CNN perspective, providing inspiration for subsequent research endeavors. The fusion of transformers and CNNs offers a flexible and diverse range of integration methods. Future research should strive to deepen the understanding of their interactions to improve design and optimization.

\subsection{Object Detection of Antenna Interference Sources}
Regularly monitoring and mitigating antenna interference sources has become one of the most critical tasks in the wireless communication field. In the past, detecting antenna interference sources mainly relied on traditional techniques such as spectrum analysis, signal recognition and positioning. However, these methods have many limitations. For example, when detection personnel identify a radio interference signal through a spectrum analyzer, they can determine only the approximate direction of the interference source based on the strength of the received signal and cannot accurately determine its position. 

\begin{figure*}[!t]
	\centering
	\includegraphics[scale=0.45]{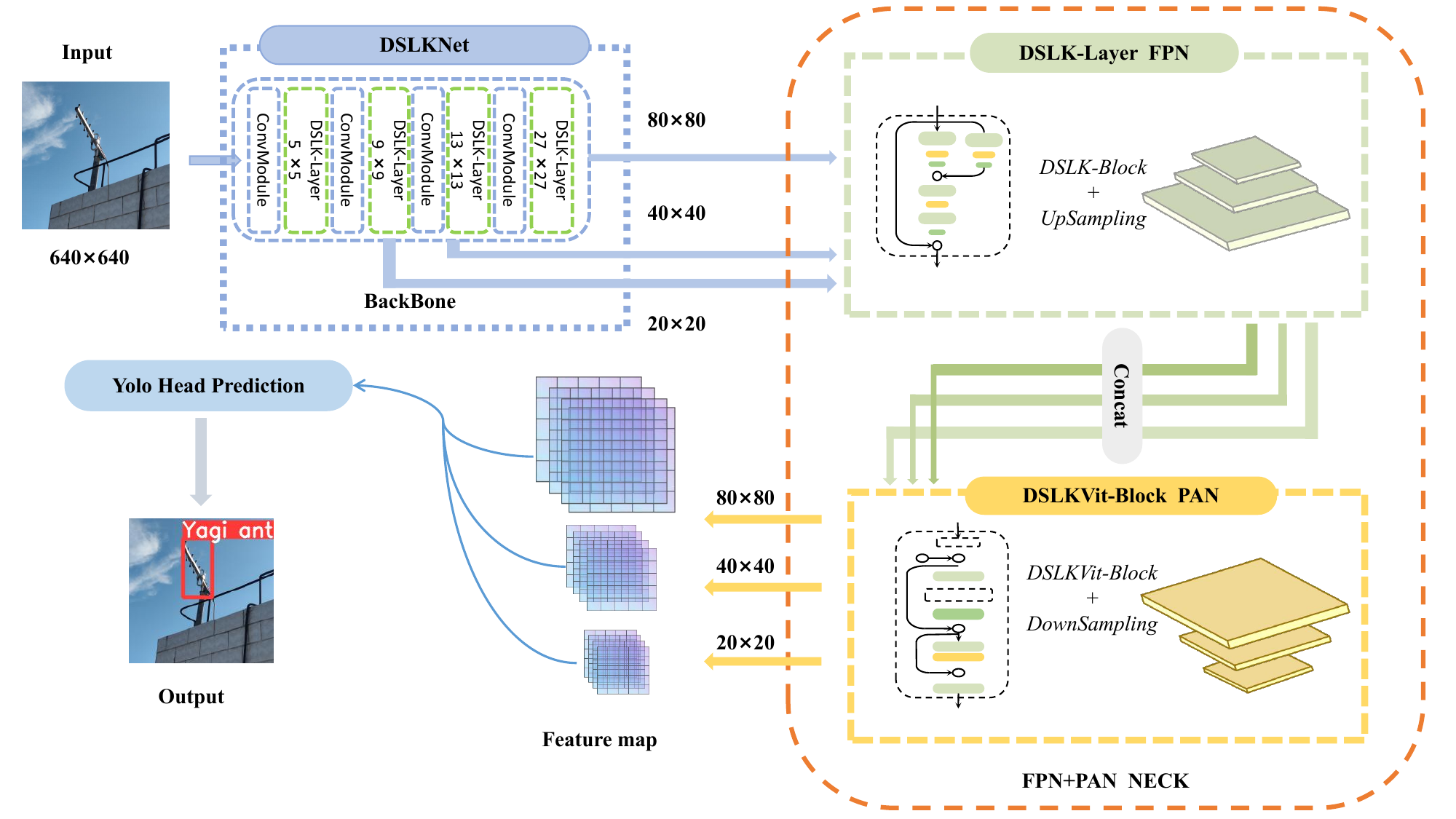}
	\caption{The YOLO-Ant model can be roughly divided into three parts: the backbone consisting of DSLKNet, the FPN+PAN structure consisting of DSLK-Layer and DSLKVit-Block forming the neck.}
  \vspace{-14pt} 
\label{figure2}
\end{figure*}

The rapid advancement of deep learning and computer vision has facilitated the successful application of object detection-assisted tasks in various industries. Examples include defect detection in industrial settings, pest/weed detection in agriculture, and vehicle and pedestrian detection in transportation \cite{guan2022lightweight}\cite{dai2022yolo}\cite{wang2022r}\cite{60376731d3485cfff1db8321}\cite{DBLP:journals/cea/DangCLL23}. These solutions provide effective ideas for our antenna interference source detection task. When investigators confirm the approximate direction of the interference source antenna through a signal receiver and spectrum analyzer, they can use drones with cameras and related object detection algorithms to replace manual accurate positioning work. Unfortunately, the field of antenna interference source detection based on object detection tasks has largely not been explored. Due to the lack of learning samples and models for related antenna interference source detection, existing detection methods are not suitable for antenna detection. Therefore, it is urgent and meaningful to create a professional dataset and train a model suitable for this detection task to address the difficulty of locating antenna interference sources in the wireless communication field.

\section{PROPOSED DETECTION FRAMEWORK}
\subsection{Overall model structure}
The overall idea for the network(Fig. \ref{figure2}) lies in the combination of a CNN and transformer, both the inductive bias ability of the convolutional operation and the ability of the transformer to extract global information, while also meeting the needs of a lightweight model with low computational complexity. YOLO-Ant adopts DSLKNet, which is composed of DSLK-Blocks, as the backbone for downsampling and feature extraction in images. In DSLKNet, four DSLK-Layers employ convolutional kernels of varying sizes to sequentially extract rich features from different receptive fields of the image. To address the challenge of detecting small objects, we incorporate the neck structures of the FPN and PAN for multiscale feature learning. On the neck component, we conducted pruning based on YOLOv5-s (detailed data provided in Section IV. EXPERIMENT). In comparison to the baseline model, the pruned neck model features an increased number of module stacks and a reduced number of channels in each module. This structural modification effectively alleviates redundancy in feature extraction, resulting in an overall model that is not only more lightweight but also attains higher detection accuracy. In this lightweight framework, we introduce the DSLKVit-Block, which is a combination of transformer and convolutional modules. Even though the transformer module has a larger parameter count and computational complexity than does the convolutional module, the final YOLO-Ant model remains lighter than the baseline model. Overall, the integration of CNN and transformer in the model is manifested as follows: (i) a transformer-like pure CNN structure is proposed using depthwise separable convolution and a large convolution kernel, effectively expanding the perceptual field of the convolution operation and extracting the target context information; (ii) the FPN structure of the pure CNN and the PAN structure designed with the transformer module are combined in a parallel structure to complement each other and thus improve the feature processing capability of the model. The following two subsections provide a detailed description of the working principles of the DSLK-Block and DSLKVit-Block within the YOLO-Ant.

\subsection{More efficient feature extraction module, DSLK-Block}
The DSLK-Block structure built with depthwise separable convolution was introduced using large convolutional kernels. The design of this structure is based on several starting points. (1) Depthwise separable convolution is used instead of conventional convolution operations to achieve model lightweighting. (2) Large convolutional kernels are used to increase the receptive fields to extract a greater amount of contextual information. Models such as RepLKNet have shown that pure convolutional networks can achieve performance comparable to that of transformer-style networks in this way. (3) Inspired by the ConvNeXt approach, DSLKBlock uses fewer normalization and activation functions, replacing the rectified linear unit (ReLU) with a Gaussian error linear unit (GELU).

However, unlike other feature extraction blocks, DSLKBlock has three places that change correspondingly with the network location of the DSLK-Layer to balance the relationships between parameter volume, computational complexity, and accuracy. (1) The size of the large convolutional kernel in the backbone changes according to the location of the DSLKBlock. The rationale behind this design primarily stems from several considerations. First, if all DSLK-Layers adopt excessively large convolutional kernels, the model will be greatly burdened in terms of parameters and detection speed. Second, small objects typically have pixels in the range of 32 $\times$ 32, roughly equivalent to 1/40 of the original image size. When the network input size is set to 640 $\times$ 640, the corresponding size of the small objects is approximately 16 $\times$ 16. To ensure that the early layers of the network can sufficiently extract feature information from small objects and prevent the loss of small object details caused by large convolutions, we employ 5 $\times$ 5 and 9 $\times$ 9 convolutional kernels in the first two layers. Finally, as the downsampling rate increases, the feature map sizes decrease. We employ larger convolutional kernels to handle larger-sized objects, while these larger kernels also provide more comprehensive contextual information for small feature maps (derived from the relationships between adequately extracted small objects and their surrounding environments). In the final model backbone, the sizes of the large convolutional kernels are set to 5 $\times$ 5, 9 $\times$ 9, 13 $\times$ 13, and 27 $\times$ 27, thereby achieving efficient detection, particularly for objects of different sizes, especially small objects. (2) For the DSLK-Layer within the model backbone, to prevent the potential loss of important information as the convolutional kernel size increases, we incorporated a parallel pathway using 3 $\times$ 3 convolutional kernels within the DSLK-Block. The output from the small kernel pathway is fused with that of the large kernel pathway using an addition operation. To ensure model lightweightness and expedite model convergence, within the neck structure, all the large convolutional kernels within the DSLK-Block were resized to 3 $\times$ 3 dimensions, while the small convolutional kernel pathway was modified to follow a conventional shortcut form. (3) Depthwise separable convolution first uses depthwise convolution to convolve each feature point within the same channel and then extracts information between different channels of the same feature point through pointwise convolution. After decomposing conventional convolution operations into these two steps, the computational cost of the convolution operation is effectively reduced. The pointwise convolution of the DSLK-Block adopts a variable factor to control the number of channels. To further balance the relationships between the model parameters, computational complexity, and accuracy, the DSLK-Block uses different pointwise convolution depths at different positions in the network and achieves good results. The final structures of DSLK-Block and DSLK-Layer are shown in Fig. \ref{figure3}. The DSLK-Block is part of the DSLK-Layer, as illustrated on the right-hand side of Fig. \ref{figure3}, and forms each CNN feature processing layer within the backbone and neck of the model. Assuming that the input feature map is denoted as $X_{i} \in R^{C_{i}\times H_{i}\times W_{i}}$(where C, H and W denote the number of channels, spatial height and width, respectively), its workflow can be represented as follows:

\begin{figure*}[!t]
	\centering
	\includegraphics[scale=0.7]{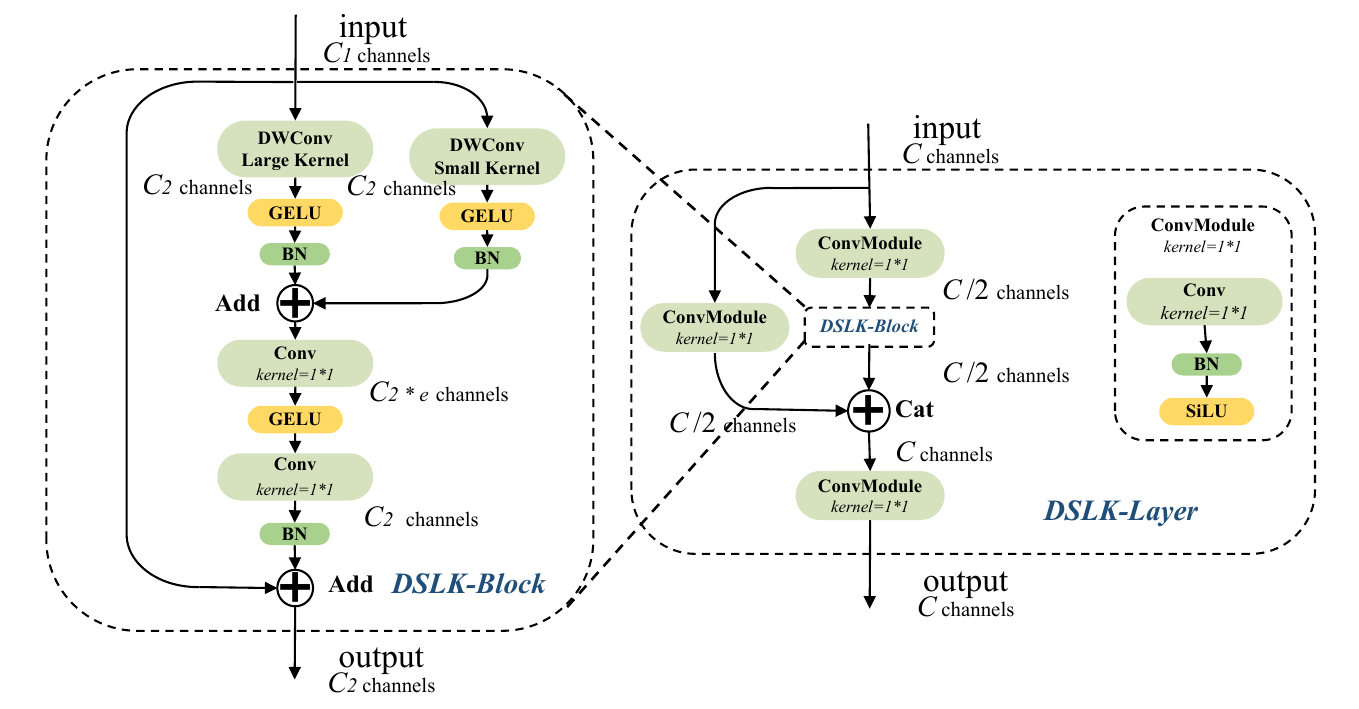}
	\caption{DSLK-Block \& DSLK-Layer}
 \vspace{-14pt} 
\label{figure3}
\end{figure*}

\begin{equation}
	\begin{aligned}
		F_{DSLK-Layer}(X_i)=f_{CBS}(1,C_i,f_{CBS}(1,C_i/2,X_i))\\
		 \otimes F_{DSLK-Block}(f_{CBS}(1,C_i/2,X_i))
	\end{aligned}
\end{equation}
where $\otimes$ represents the Concat operation and $f_{CBS}$ represents a module that sequentially undergoes convolution, normalization, and activation functions, which can be expressed using the following formula:
\begin{equation}
	\begin{aligned}
		f_{CBS}(k,c,X_i)=\rho(\sigma(f_{conv}(k,c,1,X_i)))
	\end{aligned}
\end{equation}
where $\rho(x)$ represents batch normalization and $\sigma(x)$ represents the SiLU activation function. where $f_{conv}(k,c,g,X_{i})$  represents the convolution operation, $k$ is the kernel size, $c$ is the output channel number, and $g$ represents the number of groups ($g$=1 in regular convolution, and $g=C_{i}$ in depthwise separable convolution).

We represent the workflow of the DSLK-Block using formula $F_{DSLK-Block}$:
\begin{equation}
\begin{aligned}
	F_{DSLK-Block}(X_{i})=X_{i} + f_{pw}(e,\\
	f_{dw}(K_L,X_{i}) + f_{dw}(K_S,X_{i}))
\end{aligned}
\end{equation}
where $K_L$ and $K_S$ represent the large and small convolution kernels used in the two depthwise convolution paths respectively  ($K_L$ takes values of 3, 5, 9, 13, and 27 in the model. $K_S$ takes a value of 0 or 3. If $K_L$=0, then the path becomes a normal shortcut operation). where $f_{dw}$ represents the substitution of conventional convolutions with depthwise convolution in $f_{CBS}$, as expressed by the following formula:
\begin{equation}
	f_{dw}(K,X_{i})=\rho\left(\sigma\left(f_{conv}(K,C_{i},C_{i},X_{i})\right)\right)
\end{equation}
while $f_{pw}$ represents the pointwise convolution block, as expressed by the following formula:
\begin{equation}
\begin{aligned}
	f_{pw}(e,X_{i})=\rho(f_{conv}(1,1,C_{i},\sigma(\\
	f_{conv}(1,1,e\times C_{i},X_{i}))))
\end{aligned}
\end{equation}
where $e$ represents the variable expansion coefficient, which is used to control the channel expansion and scaling factor in the pointwise convolution process.

\subsection{DSLKVit-Neck structure for efficient feature fusion}
The efficiency of transformer models relies heavily on their global attention mechanism, which differs from convolution operations in that information between feature points is calculated only within the size of the convolution kernel. Instead, transformer models consider the interactions between each feature point and all other points in the feature map. Using the transformer's self-attention mechanism to enhance the contextual information extraction of feature points is useful for dealing with problems such as target shape differences caused by multiangle shots from UAVs and interference from complex environmental backgrounds in antenna detection tasks. However, this design comes at the cost of consuming a significant amount of computational resources, making it unacceptable for lightweight models. Therefore, finding methods to efficiently utilize transformer structures while conserving resources remains a significant challenge.

\begin{figure}[!t]
	\centering
	\includegraphics[scale=0.45]{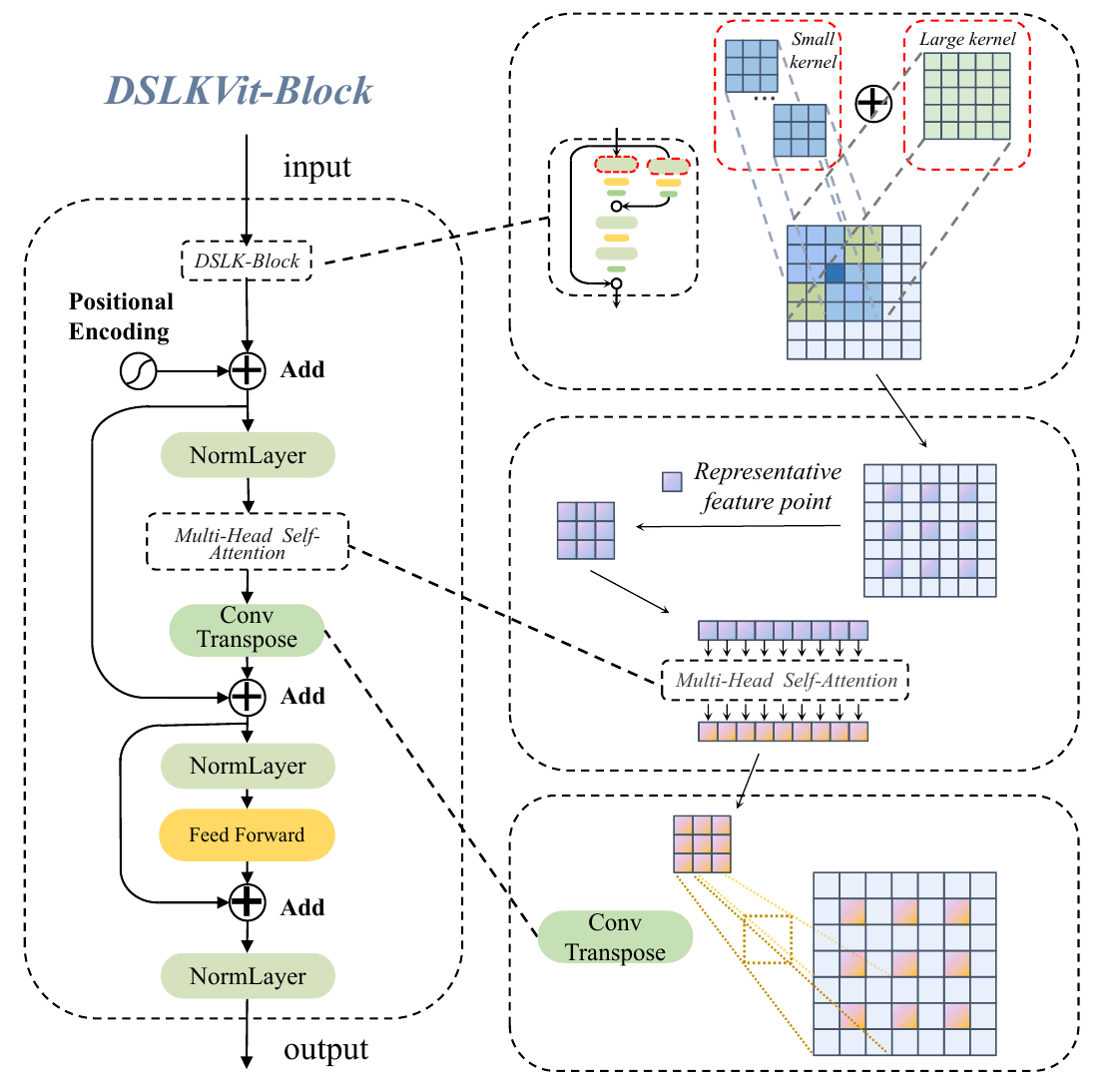}
	\caption{DSLKVit-Block}
 \vspace{-14pt} 
\label{figure4}
\end{figure}

In MobileViT, the transformer calculates information only between feature points at the same position in each patch of the feature map. In EdgeViT, a convolution operation is used to aggregate all the information within a local window of size k $\times$ k. Then, the transformer calculates the information between all the feature points that contain the aggregated information. Due to the large amount of data redundancy inherent in image data, the difference in information between adjacent pixels is often not significant, allowing this computational savings to occur. Motivated by models such as MobileViT and EdgeViT, we introduce an innovative DSLKVit-Block module, which incorporates both CNN and transformer architectures, as shown in Fig. \ref{figure4}. To reduce the computational complexity of the entire model, DSLKVit-Block calculates information only between ``representative feature point'' within small regions of the feature map instead of computing the mutual information between every feature point.

The DSLKVit-Block initially conducts feature extraction on localized regions of the feature map through convolution operations within the DSLK-Block, resulting in a set of ``representative feature points''. Each of these feature points represents the aggregated features from their respective regions. The new feature map, composed of all these ``representative feature points'', has reduced dimensions, conserving resources for subsequent self-attention computations. Assuming that the input feature map is denoted as $X_{i} \in R^{C_{i}\times H_{i}\times W_{i}}$(where C, H and W denote the number of channels, spatial height and width, respectively), we denote the process of aggregating local information in the DSLKViT-Block as follows using the formula $f_{Local}$:
\begin{equation}
	\begin{aligned}
		f_{Local}(X_i)=f_{LA}(sr, \rho(f_{PE}(X_{i}) \\
  + F_{DSLK-Block}(X_{i})))
	\end{aligned}
\end{equation}
where $\rho(x)$ represents the NormLayer operation, $f_{PE}$ represents the positional encoding achieved through convolution operations and $f_{LA}(sr,x)$ represents the operation of aggregating local information, which can be achieved using pooling layers. The $sr$ parameter indicates the local range covered by the ``representative feature points'' and is equal to $sr \times sr$.

Subsequently, the feature map undergoes multihead attention calculations, enabling the interactions among the ``representative feature points'' that aggregate information within each region, thereby acquiring rich contextual information between various regions on the original feature map. Finally, the model employs deconvolution operations to map the ``representative feature points'' back to their respective corresponding regions and enhances the network's expressive capabilities through a feed-forward network (FFN). The process of obtaining global information can be expressed using the following formula:
\begin{equation}
	\begin{aligned}
		f_{Global}(sr,X_i)=X_i + \rho(f_{LD}(sr,\\
		MHSA(f_{Local}(X_i)))
	\end{aligned}
\end{equation}
where $f_{LD}(sr,x)$ represents the operation of mapping representative feature points back to their original regions, which can be accomplished via deconvolution. $MHSA$ represents the multihead self-attention computation:
\begin{equation}
	\begin{aligned}
	\operatorname{MHSA}(X)=\operatorname{MultiHead}(Q, K, V)\\
	=Concat(head_1,head_2,...,head_h)W^{O}
	\end{aligned}
\end{equation}
where $head_i = Attention(QW_{i}^{Q},KW_{i}^{K},VW_{i}^{V})$, $W_{i}^{Q} \in R^{d_{model} \times d_k}$, $W_{i}^{K} \in R^{d_{model} \times d_k}$, $W_{i}^{V} \in R^{d_{model} \times d_v}$, and $W^{O} \in R^{hd_{v} \times d_{model}}$. In this work we employ $h=4$ parallel attention heads. For each of these we use
$d_k=d_v=d_{model}/4=C$. For single headed attention, we compute the attention function on a set of queries simultaneously, packed together into a marix $Q$. The keys and values are also packed together into matrices $K$ and $V$. We compute the matrix of outputs as follows:
\begin{equation}
	\operatorname{Attention}(Q, K, V)=\operatorname{Softmax}\left(\frac{Q K^{ T}}{\sqrt{d_k}}\right) V
\end{equation}
where $d_k$ is a scaling factor. We can obtain $Q \in R^{N \times C}$,$K \in R^{N \times C}$,and $V \in R^{N \times C}$ through a linear layer by input feature $X \in R^{C\times H \times W}$ respectively, where $N=H \times W$.

Ultimately, the entire workflow of DSLKVit-Block can be represented as follows:
\begin{equation}
	\begin{aligned}
		F_{DSLKVit}(sr,X_i)=FFN(f_{Global}(sr,(f_{Local}(X_i))))
	\end{aligned}
\end{equation}

 \vspace{-8pt} 
  
\begin{equation}
	\begin{aligned}
		FFN(X_i)=X_i + \rho(MLP(X_i))
	\end{aligned}
\end{equation}
where the $MLP$ function can be implemented using two fully connected layers. 

After finalizing the foundational DSLKVit-Block module, we proceeded to modify the model's neck structure at the macro level, aiming to harness the complementary advantages of transformers and CNNs to their fullest extent. We observed that the neck structure of YOLOv5-s adopts an FPN with a PAN architecture. In this configuration, the FPN's top-down structure merges high-level feature maps with low-level feature maps to propagate semantic information from higher levels, enhancing the representational capacity of the lower-level feature maps. Subsequently, the connection of the PAN, a bottom-up structure, aims to assist high-level feature maps in acquiring richer positional information from shallow-level feature maps. The PAN and FPN also engage in lateral connections to fuse feature maps of the same dimensions, enriching the information content. Therefore, the use of the transformer to construct PAN is logically sound. During the bottom-up process, the DSLKVit-Block gradually extracts finer features from large-sized feature maps to small-sized feature maps. 
Moreover, lateral connections are leveraged to incorporate the convolution results of the DSLK-Block to compensate for the insensitivity of the transformer structure to positional information. Furthermore, due to the computational complexity of the transformer being $O(N^2)$, applying it to feature maps of size 80 $\times$ 80 would results in a prohibitively high computational cost. Therefore, we deployed the DSLKVit-Block only in the smallest two feature layers (40 $\times$ 40 and 20 $\times$ 20). 

The DSLKVit-Neck structure constructed from the above design incorporates the complementary concepts of CNNs and transformers at both the microlevel (within DSLKVit-Block) and macrolevel (the entire model's neck structure).

\section{EXPERIMENT}
\subsection{Experimental Description}
\noindent {\bf{Dataset:}}
\begin{enumerate}{}{}
	\item{Antenna interference source dataset: In the antenna interference source detection task proposed in this article, the methods and datasets in the relevant field are still lacking. Therefore, our first priority is to create the first dataset specifically for antenna interference source detection. In the daily interference source investigation and cleaning activities of the communication bureau, we identified three common antenna interference sources that significantly impact communication: the Yagi antenna, plate log antenna, and patch antenna, as shown in Fig. \ref{dataset_ant} a-c. The three antennas have different shapes, bringing multiple challenges to the detection task. For example, the Yagi antenna is easily confused with a complex fence background when it is installed on a balcony, as shown in Fig. \ref{dataset_ant}d. The plate log antenna has a flat and wide shape that can be recognized from the front well. However, the angle from the side greatly reduces the recognition rate, and different angles create problems with significant interclass differences, as shown in Fig. \ref{dataset_ant} e. The patch antenna is the most difficult to recognize because it is merely a monochromatic rectangular object accompanied by a lengthy wire, resulting in a lack of intricate features available for the model to learn from. Additionally, when the patch antenna is wrapped around a pole and placed high up, the object captured by aerial photography is small and difficult to distinguish in the image, as shown in Fig. \ref{dataset_ant}f. Moreover, we simulate different angles, distances, and lighting effects to enrich the dataset when drones take pictures of the antennas. Based on the collected images of the three types of antenna interference sources, we used common data augmentation techniques such as rotation, flipping, and image color space changes to expand the original images. Finally, the dataset was divided into training images (1777) and validation images (449) at an 8:2 ratio and labeled by professionals.}
    \begin{figure}[!t]
		\centering
		\includegraphics[scale=0.55]{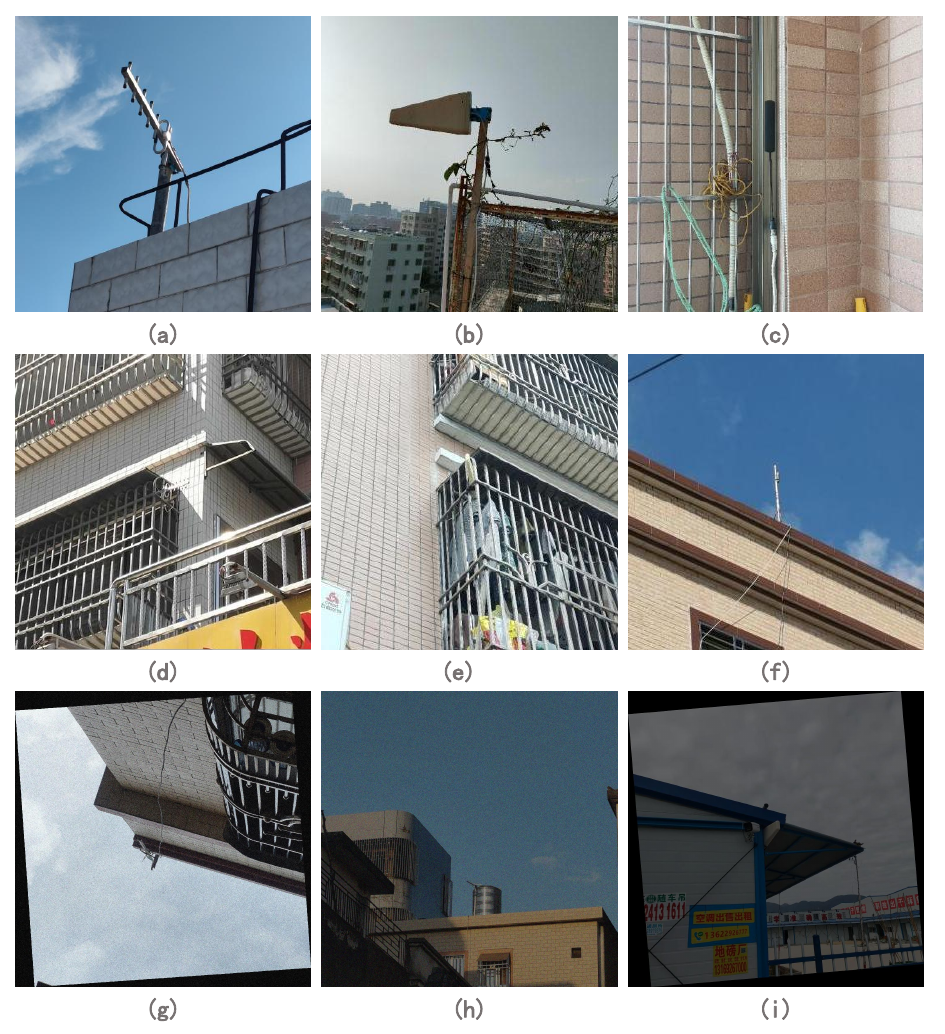}
		\caption{Antenna Interference Source Dataset}
    \vspace{-16pt} 
    \label{dataset_ant}
    \end{figure}
	\item{COCO: The Common Objects in Context (COCO) dataset is widely used in the computer vision community for benchmarking object detection and segmentation algorithms. It is designed for use in object detection, segmentation, and captioning. The dataset contains more than 330,000 images with more than 2.5 million object instances labeled across 80 different object categories. Additionally, the dataset includes 91 different categories of “stuff” or background categories, such as sky, grass, and water. The dataset also includes annotations for object segmentation masks, object bounding boxes, and keypoint annotations.}
	\item{VisDrone: The VisDrone dataset is a large-scale benchmark dataset for visually understanding aerial scenes. The dataset consists of more than 6,000 video clips and more than 2 million images captured by various types of drones in different locations and scenarios, covering a wide range of aspects of aerial vision. The dataset contains rich annotations for object detection, tracking, counting, and image quality assessment, which are important for both academic research and industrial applications.}
\end{enumerate}

\noindent {\bf{Experimental Setting:}} Experimental Configuration: All the experiments were executed on the Linux operating system, an NVIDIA GTX3060 GPU with 12 GB of  memory, and data training and object detection were performed in PyTorch 1.12.0, torchvision0.13.0, and CUDA10.2 environments. On two public datasets, COCOtrain2017 and VisDrone2019-DET-train, we conducted training and testing on the COCOval2017 and VisDrone2019-DET-val validation sets, respectively. In the experiments with the COCO dataset, YOLO-Ant was trained without the use of pretrained weights, and the model was tested using the weights that achieved the highest mAP.5:.95 after training to full convergence. For the Antenna dataset and the Visdrone dataset, training was more challenging due to the smaller number of samples and the difficulty posed by small targets. To improve the model fit and enhance the accuracy, all the models were trained using pretrained weights (models pretrained on the COCO dataset). To ensure a fair comparison, all the experiments involving model training, testing, and other evaluations are conducted using images of size 640x640 as the input. All pretrained weights for comparative models in the experiments, as well as settings for hyperparameters such as learning rates and optimizers, were sourced from the official project or MMDetection's official documentation.

\noindent {\bf{Evaluation Criteria:}} In the object detection field, the following metrics are used to judge the detection accuracy of a model \cite{guoudtiri}: accuracy (P), recall (R), average precision (AP), mean AP (mAP), and intersection over union (IoU). The accuracy is the proportion of positive cases predicted by the model that are actually positive. Recall is the proportion of all true positive cases that the model correctly predicts as positive cases. AP is the weighted average of accuracy and recall. The mAP is the mean AP value for different categories. The IoU is the ratio of the intersection area of the predicted frame to the true frame to the concatenated area and is usually used to evaluate the localization accuracy of an object detection model. mAP.5 indicates the mAP when the IoU threshold is 0.5; mAP.5:.95 evaluates the performance of the algorithm over a range of IoU thresholds from 0.5 to 0.95, as the detection model performance under different IoU thresholds is considered, allowing a more comprehensive assessment of the detection and localization capabilities of the model.

\vspace{-5pt}

\begin{equation}
	\mathrm{P}=\frac{T_P}{T_P+F_P}
\end{equation}

\vspace{-5pt}

\begin{equation}
    \mathrm{R}=\frac{T_P}{T_P+F_N}
\end{equation}

\vspace{-5pt}

\begin{equation}
    \mathrm{AP}=\int_0^1 P(R) d R \\
\end{equation}

\vspace{-5pt}

\begin{equation}
	\mathrm{mAP}=\frac{\sum_{i=1}^N A P(i)}{N}
\end{equation}

In the formula, TP (true positive) is the correct detection result, FP (false positive) is the wrong detection result, and FN (false negative) is the wrong undetected result, which means missed detection. N indicates the number of detection task categories.

\begin{table*}[!t]
	\centering
	\caption{The performance of different models on the Antenna dataset}
	\label{tab:4}  
	\begin{tabular}{ccccccccc}
		\toprule 
		model & Year & Param. & GFLOPs & mAP.5 & mAP.5:.95 & Yagi Antenna & Plate Log Antenna  & Patch Antenna \\
		      &      &        &        &       &           & mAP.5 & mAP.5  & mAP.5 \\
		\hline 
		RetinaNet(ResNet50)\cite{lin2017focal} & ICCV'2017 &37.97M & 191.43 & 0.477 & 0.237 & - & - & -\\
		YOLOv3-tiny\cite{redmon2018yolov3} &'2018 &8.85M & 13.17 & 0.32 & 0.156 & 0.288 & 0.645 & 0.0274 \\
        FCOS(ResNet50)\cite{tian2019fcos} & ICCV'2019 &32.12M & 60.59 & 0.533 & 0.264 & - & - & -\\
		YOLOv4-tiny\cite{bochkovskiy2020yolov4} & '2020 &5.88M & 16.18 & 0.28 & 0.134 & - & - & -\\
            YOLOv5-s\cite{glenn_jocher_2021_5563715} & '2020 &7.23M & 16.49 & 0.619 & 0.339 & - & - & - \\
		YOLOx-s\cite{yolox2021} & CVPR'2021 &8.94M & 26.76 & 0.579 & 0.350 & 0.650 & 0.781 & 0.306 \\
		YOLOv7-tiny\cite{wang2022yolov7} & CVPR'2023 &6.23M & 13.86 & 0.42 & 0.217 & 0.476 & 0.642 & 0.141 \\
		EfficientDet-D1\cite{tan2020efficientdet} & CVPR'2020 &6.56M & 11.51 & 0.267 & 0.156 & - & - & - \\		
		TOOD(ResNet50)\cite{feng2021tood} & ICCV'2021 &32.02M & 60.79 & 0.499 & 0.265 &  &  & \\
		PVT-tiny\cite{wang2021pyramid}  & ICCV'2021 &23.0M & 50.94 & 0.472 & 0.246 & - & - & - \\
		DAB-DETR(ResNet50)\cite{liu2022dabdetr}& ICLR'2022 &43.70M & 33.90 & 0.531 & 0.227& - & - & -\\
           YOLOv8-s\cite{Jocher_YOLO_by_Ultralytics_2023}&'2023 &11.2M & 28.4 & 0.548 & 0.316& - & - & -\\
		DINO(ResNet50)\cite{zhang2022dino}& ICLR'2023 &47.54M & 93.77 & 0.595 & 0.331 & - & - & - \\
          Mask R-CNN(ConvNeXt-V2-B\cite{Woo2023ConvNeXtV2})&  '2023 &110.59M &  198.56 & 0.642 & 0.348 & - & - & - \\
            RT-DETR(ResNet18)\cite{lv2023detrs}& '2023 &20M & 60 & 0.653 & 0.362 & - & - & - \\	
		\hline 
		ours & '2023 &6.13M & 16.18 & 0.692 & 0.374 & 0.739 & 0.825 & 0.512 \\
		\bottomrule 
	\end{tabular}
\label{baseline}
\end{table*}

In addition to model accuracy, the number of parameters and giga floating-point operations per second (GFLOPs) are crucial considerations for determining the lightweight nature of a model. FLOPs indicate the number of floating-point operations performed by a model in a single forward propagation, serving as a measure of its computational complexity. This metric enables the comparison of computational overhead across different models. GFLOPS represents the billion floating-point operations per second, with 1 GFLOPs equaling 1,000 MFLOPS. In general, the more complex the model structure is, the higher the number of FLOPs. When designing an object detection model, a trade-off between accuracy and computational complexity is needed to achieve better detection performance and higher computational efficiency by making a corresponding choice according to the specific task requirements.

\subsection{Experimental Results and Analysis}

\noindent {\bf{Baseline Model Selection:}} 
Starting from the practical task of antenna interference source detection, we select classic or cutting-edge algorithms (with a focus on lightweight methods) for comparison on our antenna dataset. The selection process encompasses considerations from three main aspects: model parameter count, computational complexity, and detection accuracy. It is evident from Table \ref{baseline} that with the increased popularity of transformers in the computer vision domain in recent years, most detection methods have become closely associated with them. However, this trend has led to increasingly larger models, rendering them unsuitable for lightweight real-world applications. In terms of detection accuracy, several cutting-edge transformer models such as PVT-tiny\cite{wang2021pyramid} and DAB-DETR, not only fail to meet the lightweight criteria but also exhibit detection performances inferior to those of traditional single-stage detectors, such as FCOS\cite{tian2019fcos} and YOLO. The reason behind this phenomenon lies in the inherent limitations of transformer mechanisms for small object detection.

Taking a holistic view, among this group of the lightest models (enclosed by the red box in Fig. \ref{zhexian}(c)), there is a significant disparity in model accuracy, with most mAP.5 values falling below 0.4. In contrast, among some larger models (enclosed by the green box in Fig. \ref{zhexian}(c)), the accuracy on the antenna dataset is higher and more consistent, with mAP.5 values generally hovering at approximately 0.5. We observed that lightweight models such as YOLOv4-tiny and EfficientDet-D1 are constrained by their simplistic network structures and computational limitations, making it challenging for them to achieve outstanding performance on antenna datasets characterized by small targets and complex backgrounds. On the other hand, larger models have better feature processing capabilities, resulting in overall superior accuracy. However, to meet the demand for deploying models on low-level computing platforms, lightweight models remain a crucial focus.
From this perspective, even though RT-DETR\cite{lv2023detrs} achieves the highest accuracy among numerous comparative models, its computational complexity remains significantly higher than that of YOLO models when using a ResNet18 backbone. Additionally, due to its complex transformer architecture, the model deviates from the lightweight direction. While YOLOv7-tiny boasts lower computational complexity and model parameter count while achieving an mAP.5 above 0.4, its detection accuracy still lags behind that of YOLOv5-s. Therefore, YOLOv5-s is appropriately selected as the baseline model for subsequent experiments.

\begin{table*}[!t]
	\renewcommand{\arraystretch}{1.5}
	\centering
	\caption{Baseline model pruning}
	\label{table1}
	\begin{tabular}{|cc|cc|cc|}
		\hline
		\multicolumn{2}{|c|}{}                         & \multicolumn{2}{c|}{YOLOv5-s}                                           & \multicolumn{2}{c|}{YOLOv5s -lite}                                       \\ \hline
		\multicolumn{1}{|c|}{Layer Name} & Output size & \multicolumn{1}{c|}{modules{[}in channels,out channels{]}$\times$n} & params  & \multicolumn{1}{c|}{modules{[}in channels,out channels{]}$\times$n} & params \\ \hline
		\multicolumn{1}{|c|}{stage 17}   & 80$\times$80       & \multicolumn{1}{c|}{C3{[}256,128{]}$\times$1}                       & 90880   & \multicolumn{1}{c|}{C3{[}256,128{]}$\times$3}                       & 173312 \\
		\multicolumn{1}{|c|}{stage 18}   & 80$\times$80       & \multicolumn{1}{c|}{Conv{[}128,128{]}$\times$1}                     & 147712  & \multicolumn{1}{c|}{Conv{[}128,128{]}$\times$1}                     & 147712 \\
		\multicolumn{1}{|c|}{stage 20}   & 40$\times$40       & \multicolumn{1}{c|}{C3{[}256,256{]}$\times$1}                       & 296448  & \multicolumn{1}{c|}{C3{[}256,128{]}$\times$3}                       & 173312 \\
		\multicolumn{1}{|c|}{stage 21}   & 40$\times$40       & \multicolumn{1}{c|}{Conv{[}256,256{]}$\times$1}                     & 590336  & \multicolumn{1}{c|}{Conv{[}256,128{]}$\times$1}                     & 147712 \\
		\multicolumn{1}{|c|}{stage 23}   & 20$\times$20       & \multicolumn{1}{c|}{C3{[}512,512{]}$\times$1}                       & 1182720 & \multicolumn{1}{c|}{C3{[}256,128{]}$\times$3}                       & 173312 \\ \hline
		\multicolumn{2}{|c|}{GFLOPs}                   & \multicolumn{2}{c|}{16.5}                                              & \multicolumn{2}{c|}{15.7}                                             \\ \hline
		\multicolumn{2}{|c|}{Parms(all)}               & \multicolumn{2}{c|}{7235389}                                           & \multicolumn{2}{c|}{5398845}                                          \\ \hline
            \multicolumn{2}{|c|}{mAP.5}               & \multicolumn{2}{c|}{0.619}                                           & \multicolumn{2}{c|}{0.640}                                          \\ \hline
	\end{tabular}
\label{tab:1}
\vspace{-8pt}
\end{table*}

\begin{figure*}[!t]
	\centering
	\includegraphics[scale=0.5]{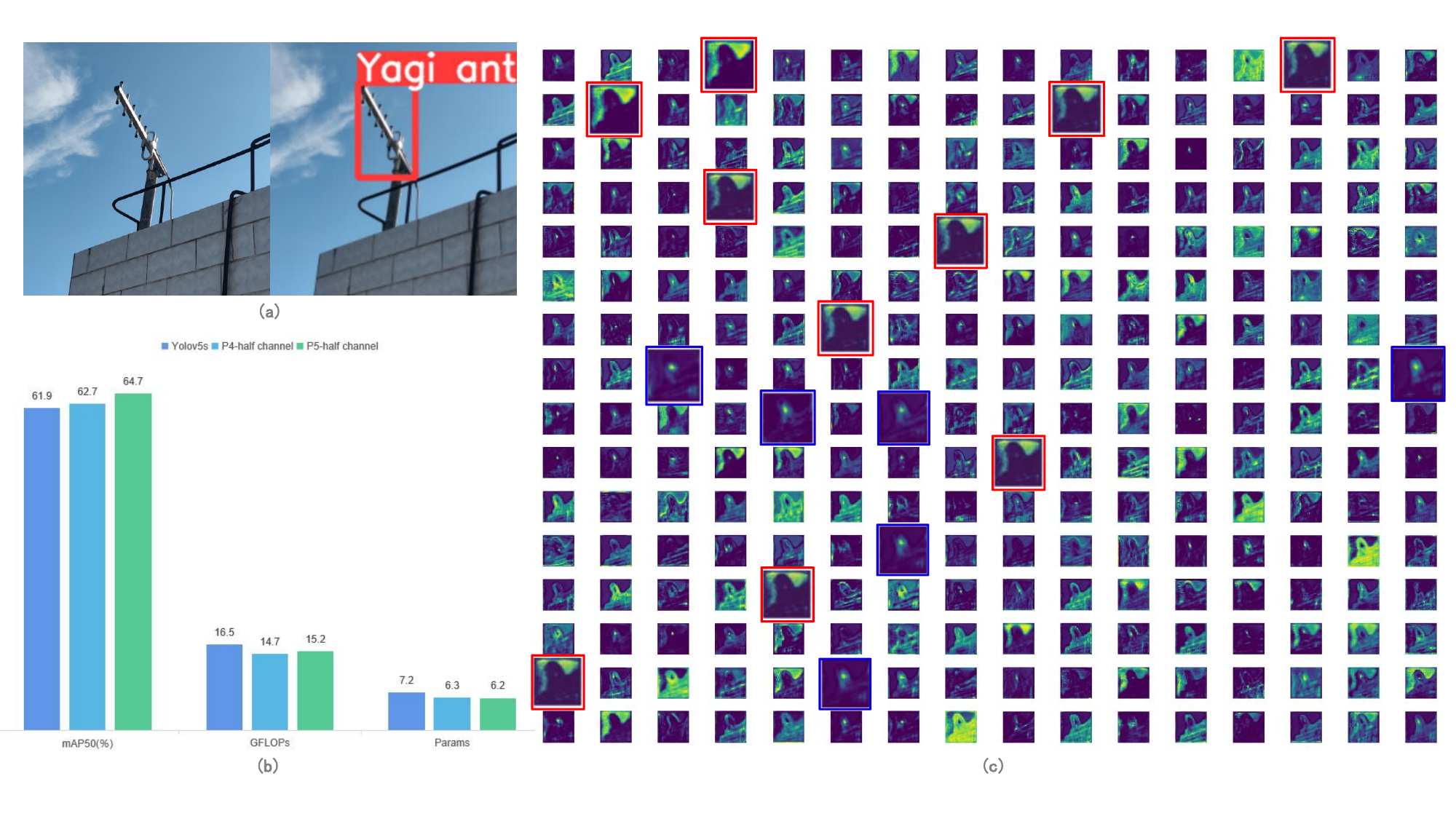}
   \vspace{-10pt} 
	\caption{(a) shows an example of an input antenna image and detection result, (b) compares the performance results of the baseline model after channel pruning on the neck, and (c) shows the visualization results of the P4 feature layer of the baseline model were presented. These visualizations unveiled the presence of a considerable degree of information redundancy among the 256 channels. This redundancy was notably characterized by a pronounced repetition of feature maps. To illustrate this phenomenon, we have highlighted two specific groups as exemplary instances.}
   \vspace{-5pt} 
\label{figure}
\end{figure*}

\begin{figure*}[!t]
	\centering
	\includegraphics[scale=0.35]{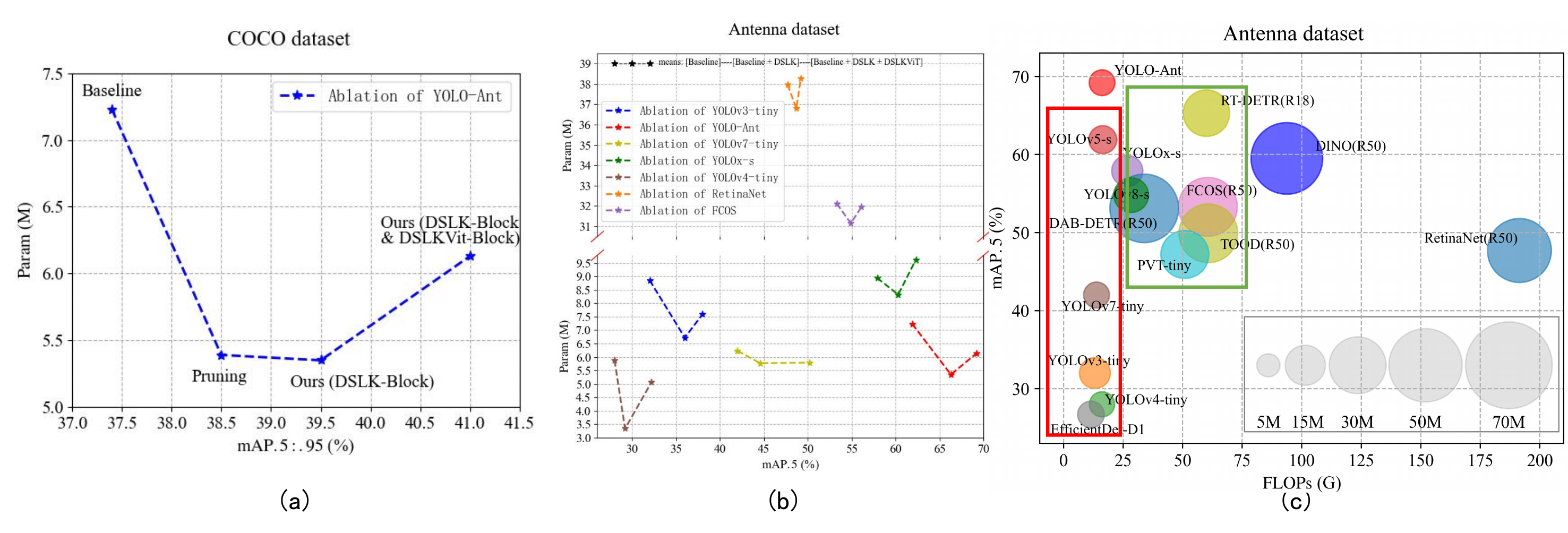}
	\caption{Figure (a) shows the corresponding changes in accuracy and number of parameters during the ablation experiment of YOLO-Ant on the COCO dataset. Figure (b) shows the performance of two modules, DSLK-Block and DSLKVit-Block, retrofitting different models on the antenna dataset. Figure (c) shows the performance of different models on the antenna dataset in three dimensions: accuracy, computational complexity and number of parameters. The models enclosed by the green box generally exhibit higher accuracy compared to those enclosed by the red box but deviate from the lightweight design principle. }
    \vspace{-5pt} 
\label{zhexian}
\end{figure*}

\noindent {\bf{Baseline Model pruning:}} 
In the neck structure of the YOLOv5-s model, the C3 module and convolutional layers exhibit a wide and shallow shape (deeper channel counts and fewer convolution module stackings) as the downsampling factor increases. Specifically, when the 640 $\times$ 640 input image is downsampled to 80 $\times$ 80 (P3), 40 $\times$ 40 (P4), and 20 $\times$ 20 (P5) in the neck structure, the number of stacked C3 modules is 1, and the channel numbers of the C3 modules and the convolutional layers in the three feature maps are 128, 256, and 512, respectively. When training YOLOv5-s on the antenna dataset, the visualization of the outputs of the P3, P4, and P5 feature layers revealed that increasing the number of channels in the neck section did not necessarily imply the extraction of richer features. As shown in Fig. \ref{figure}(c), among the 256 channels in the P4 feature map, there were varying numbers of highly redundant patterns, and this redundancy was even more prevalent in the 512 channels of the P5 feature map. These redundant feature maps impeded the model's detection speed and, to some extent, reduced the model's detection accuracy, particularly when these highly repetitive features were ineffective. Therefore, we conducted experiments by reducing the output channels of the C3 modules and convolutional layers corresponding to the P4 and P5 feature layers by half. The results demonstrated that this pruning operation not only made the model lighter but also improved the detection accuracy, as shown in Fig. \ref{figure}(b). We reasonably conclude that widening the model by increasing only the number of channels is not effective for feature extraction and fusion on the antenna dataset; it increases the number of model parameters, reduces the computational speed, and even brings negative gains.

Based on this discovery, we reduced the channel dimensions of the P3, P4, and P5 to 128. While making the model ``narrower'' in terms of its width, we enhanced its learning capability and robustness by increasing the number of stacked C3 layers. After pruning the model's neck structure into a ``narrow and deep'' configuration, the overall model not only became more lightweight but also achieved even higher detection accuracy, as shown in Table \ref{tab:1}.

\begin{table*}[!t]
	\centering
	\caption{Ablation Experiment on the COCO Dataset}
	\label{tab:2}  
	\begin{tabular}{cccccccc}
		\toprule 
		model & Param. & GFLOPs & mAP.5 & mAP.5:.95 & mAP.5(small) & mAP.5(medium) & mAP.5(large)\\
		\hline 
		YOLOv5-s\cite{glenn_jocher_2021_5563715} & 7.23M & 16.49 & 0.572 & 0.374 & 0.212 & 0.423 & 0.490 \\
		YOLOv5s-pruning & 5.39M & 15.67 & 0.570 & 0.385 & 0.222 & 0.430 & 0.498 \\
		ours(DSLK-Block) & 5.35M & 14.97 & 0.584 & 0.395 & 0.236 & 0.441 & 0.508 \\
		ours(DSLK-Block+DSLKVit-Block) & 6.13M & 16.18 & 0.599 & 0.410 & 0.245 & 0.455 & 0.535 \\
		\bottomrule 
	\end{tabular}
\end{table*}

\begin{table*}[!t]
	\centering
	\caption{Ablation Experiment on the Antenna Dataset }
	\label{tab:3}  
	\begin{tabular}{cccccccc}
		\toprule 
		model & Param. & GFLOPs & mAP.5 & mAP.5:.95 & Yagi Antenna  & Plate Log Antenna  & Patch Antenna \\
		 &  &  &  & & mAP.5 &  mAP.5 & mAP.5\\
		\hline 
		YOLOv5-s\cite{glenn_jocher_2021_5563715} & 7.23M & 16.49 & 0.619 & 0.339 & 0.682 & 0.794 & 0.383 \\
		YOLOv5s-pruning & 5.39M & 15.67 & 0.640 & 0.356 & 0.666 & 0.804 & 0.451 \\
		ours(DSLK-Block) & 5.35M & 14.97 & 0.663 & 0.363 & 0.727 & 0.793 & 0.468 \\
		ours(DSLK-Block+DSLKVit-Block) & 6.13M & 16.18 & 0.692 & 0.374 & 0.739 & 0.825 & 0.512 \\
		\bottomrule 
	\end{tabular}
 \vspace{-12pt}
\end{table*}

\begin{table*}[!t]
	\centering
	\caption{Validity of the proposed module on other models (on Antenna Dataset)}
	\label{tab:4}  
	\begin{tabular}{ccccccccc}
		\toprule 
		model & Year & Param. & GFLOPs & mAP.5 & mAP.5:.95 & Yagi Antenna & Plate Log Antenna  & Patch Antenna \\
		      &      &        &        &       &           & mAP.5 & mAP.5  & mAP.5 \\
		\hline 
		RetinaNet(ResNet50)\cite{lin2017focal} & ICCV'2017 &37.97M & 191.43 & 0.477 & 0.237 & - & - & -\\
		+DSLK-Block &  &36.80M& 184.92 & 0.487 & 0.241 &-  &-  &- \\
		+DSLK-Block+DSLKVit-Block &  &38.26M & 187.85 & 0.492 & 0.243 & - & - &- \\      
		\hline 
		YOLOv3-tiny\cite{redmon2018yolov3} &'2018 &8.85M & 13.17 & 0.32 & 0.156 & 0.288 & 0.645 & 0.0274 \\
		+DSLK-Block & &6.72M & 8.94 & 0.36 & 0.145 & 0.414 & 0.596 & 0.0682 \\
		+DSLK-Block+DSLKVit-Block & &7.58M & 11.45 & 0.38 & 0.17 & 0.37 & 0.609 & 0.16 \\
    	\hline 
        FCOS(ResNet50)\cite{tian2019fcos} & ICCV'2019 &32.12M & 60.59 & 0.533 & 0.264 & - & - & -\\
        +DSLK-Block &  &31.15M & 58.49 & 0.549 & 0.268 &- & - & - \\
        +DSLK-Block+DSLKVit-Block &  & 31.94M & 58.49 & 0.561 & 0.268 & - & - &- \\		
		\hline 
		YOLOv4-tiny\cite{bochkovskiy2020yolov4} & '2020 &5.88M & 16.18 & 0.28 & 0.134 & - & - & -\\
		+DSLK-Block &  & 3.34M & 14.53 & 0.292 & 0.145 &-  &-  &- \\
		+DSLK-Block+DSLKVit-Block & & 5.06M & 15.76 & 0.322 & 0.156 & - & - &- \\
		\hline 
		YOLOx-s\cite{yolox2021} & CVPR'2021 &8.94M & 26.76 & 0.579 & 0.350 & 0.650 & 0.781 & 0.306 \\
		+DSLK-Block &  &8.32M & 26.62 & 0.603 & 0.334 & 0.679 & 0.789 & 0.341 \\
		+DSLK-Block+DSLKVit-Block &  &9.61M & 27.81 & 0.623 & 0.350 & 0.673 & 0.763 & 0.434 \\
		\hline 
		YOLOv7-tiny\cite{wang2022yolov7} & CVPR'2023 &6.23M & 13.86 & 0.42 & 0.217 & 0.476 & 0.642 & 0.141 \\
		+DSLK-Block &  &5.77M & 12.93 & 0.446 & 0.227 & 0.512 & 0.696 & 0.13 \\
		+DSLK-Block+DSLKVit-Block &  &5.79M & 12.89 & 0.502 & 0.255 & 0.594 & 0.706 & 0.206 \\
		\bottomrule 
	\end{tabular}
\end{table*}

\noindent {\bf{Ablation Experiment:}} 
As shown in Table \ref{tab:2} and Fig. \ref{zhexian}(a),  the baseline model was tested via ablation experiments on the COCO dataset. After pruning YOLOv5-s, both the number of parameters and the computational complexity were effectively reduced. Simultaneously, all the detection indices are comparable to or even surpass the original version, among which the mAP.5:.95 improved significantly, which shows that the structural pruning operations discussed in Chapter III are reasonable. On this basis, the model adds the feature processing module DSLK-Block to replace the C3 structure of the original model, which further reduces the number of parameters by 26\% compared to the original YOLOv5-s. The computational complexity is also reduced by 1.52 GFLOPs. In terms of accuracy, this version of the model shows a more significant improvement in both mAP.5 and mAP.5:.95, with 11.3\% and 4.26\% improvements in small and medium-sized targets, respectively; these findings show that DSLK-Block plays a significant role in small object detection. Furthermore, when we constructed the PAN structure in the neck using the transformer module based on DSLK-Block, the final model was obtained. Due to the balanced relationship between computing resources and accuracy improvement, DSLKVit-Block is applied to only two feature layers of smaller sizes, P4 and P5, corresponding to medium- and large-sized objects, respectively, in the detection task. The final model has an increase in the number of parameters and the complexity of operations compared to the version using only the DSLK-Block, but both are lower than YOLOv5-s. The experimental results also show that, compared to YOLOv5-s, the detection accuracy of large targets is improved by 9.2\%, which leads to a 9.6\% improvement in mAP.5:.95; this is a more significant effect, while other indicators are also improved by approximately 0.03. 

Table \ref{tab:3} shows the ablation experiments of the proposed model on the antenna interference source dataset. The overall  performance improvement is similar to that of the ablation experiments on the COCO dataset. The pruned version achieves comparable levels of all the metrics compared to the original version; both the version with only DSLK-Block and the final model show significant improvements compared to YOLOv5-s, and the final model achieves the best performance in all the metrics. Notably, the version using only the DSLK-Block has the most significant improvement in Yagi antenna detection compared to the other two antennas, while the final version using the DSLKVit-Block has a more significant improvement in the plate log antenna and patch antenna. We believe that the efficient feature extraction capability of DSLK-Block plays a crucial role due to the more complex shape of the Yagi antenna. The plate log antenna and patch antenna are more fixed in shape and color, and simple local convolution cannot extract more effective information; therefore, a transformer, a self-attention mechanism that focuses more on global information, can effectively extract the feature information of the target and background and combine the target with the surrounding environment information, making the improvement of detection accuracy more obvious.

\vspace{-1pt}

In general, combining the ablation experiments performed by the model on the COCO dataset and the antenna dataset, we can conclude the following. a. Pruning on the neck structure based on the original version can effectively reduce the model redundancy and still ensure accuracy while effectively reducing the number of parameters and computational complexity. b. Our proposed DSLK-Block can effectively improve the feature extraction capability of models for small-sized targets, as well as detect complex environments, and is lightweight. c. The DSLKVit-Block employed in the final model offers a significant advantage in effectively utilizing global information, leading to improvements in accuracy across all aspects. Since this structure acts on the output layer for larger size object detection, it makes the model improvement for such objects more obvious and directly leads to a significant improvement in the mAP.5:.95 metric.

\begin{table*}[!t]
	\centering
	\caption{Comparative Experiments on the COCO Dataset}
	\label{tab:5}  
	\begin{tabular}{cccccccc}
		\toprule 
		model & Param. & GFLOPs & mAP.5 & mAP.5:.95 & mAP.5(small) & mAP.5(medium) & mAP.5(large)\\
		\hline 
		EfficientDet-D1\cite{tan2020efficientdet} & 6.6M & 11.51 & 0.586 & 0.396 & 0.179 & 0.443 & 0.560 \\	
		YOLOv5-s\cite{glenn_jocher_2021_5563715} & 7.2M & 16.5 & 0.568 & 0.374 & - & - & - \\
		YOLOv6-tiny\cite{li2022yolov6} & 15.0M & 36.7 & 0.566 & 0.403 & - & - & - \\
		DAMO-YOLO-tiny\cite{damoyolo} & 8.5M & 18.1 & 0.580 & 0.418 & 0.230 & 0.461 & 0.585 \\
		PPYOLOE-s \cite{xu2022ppyoloe}& 7.9M & 17.4 & 0.605 & 0.430 & 0.232 & 0.464 & 0.569 \\		
		YOLOv7-tiny \cite{wang2022yolov7}& 6.2M & 13.7 & 0.567 & 0.387 & 0.188 & 0.424 & 0.519 \\
		Mobile-Former-508M(RetinaNet) \cite{chen2022mobile}& 8.4M & - & 0.583 & 0.380 & 0.229 & 0.412 & 0.497 \\
		EdgeViT-XXS(RetinaNet)\cite{pan2022edgevits} & 13.1M & - & 0.590 & 0.387 & 0.224 & 0.420 & 0.516 \\
		PVT-tiny(RetinaNet)\cite{wang2021pyramid}& 23.0M & 50.94 & 0.569 & 0.367 & 0.226 & 0.388 & 0.500 \\
		ConT-m-tiny(RetinaNet)\cite{yan2021contnet} & 27.0M & 217.2 & 0.581 & 0.379 & 0.230 & 0.406 & 0.504 \\
		\hline 
		our & 6.1M & 16.2 & 0.599 & 0.410 & 0.245 & 0.455 & 0.535 \\
		\bottomrule 
	\end{tabular}
 \vspace{-5pt}
\end{table*}

\begin{table}[!t]
	\centering
	\caption{Comparative Experiments on the VisDrone Dataset}
	\label{tab:6}  
	\begin{tabular}{ccccc}
		\toprule 
		model & Param. & GFLOPs & mAP.5 & mAP.5:.95\\
		\hline 
		FCOS \cite{tian2019fcos}&- & -& 0.258 & 0.142 \\
		VFNet \cite{zhang2020varifocalnet}& - & -& 0.288 & 0.168 \\
		TOOD \cite{feng2021tood}&- & -& 0.294 & 0.181 \\
		RetinaNet(ResNet50) \cite{lin2017focal}& 37.97M & 191.43 & 0.214 & 0.118  \\
		YOLOx-s \cite{yolox2021}& 8.97M & 26.93 & 0.254 & 0.133 \\
		YOLOv5-s \cite{glenn_jocher_2021_5563715}& 7.23M & 16.5 & 0.284 & 0.154 \\
		YOLOv7-tiny \cite{wang2022yolov7}& 6.23M & 13.86 & 0.298 & 0.153 \\
		\hline 
		ours & 6.13M & 16.18 & 0.295 & 0.163 \\
		\bottomrule 
	\end{tabular}
\vspace{-12pt}
\end{table}

\begin{figure*}[!t]
	\centering
	\includegraphics[scale=0.66]{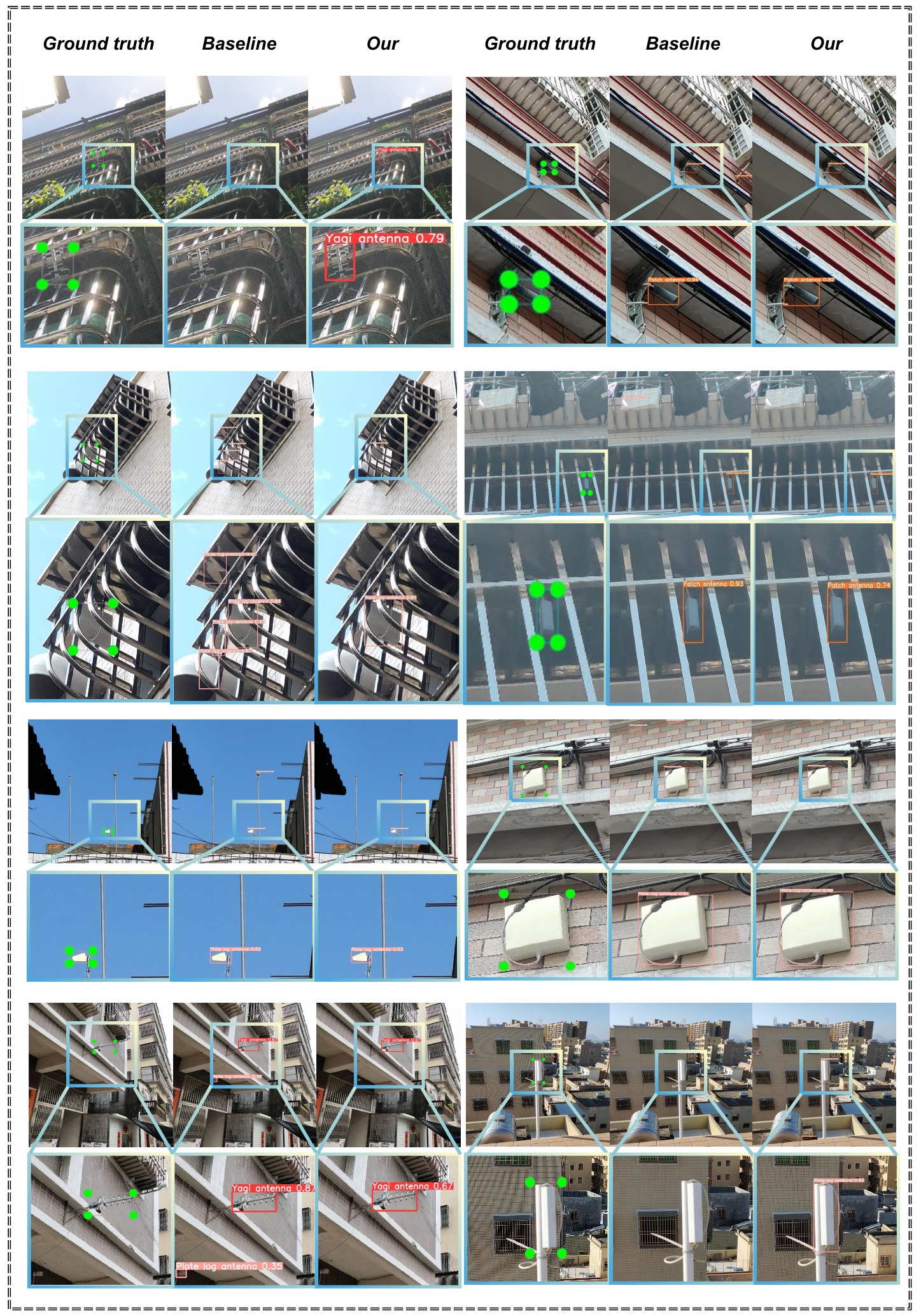}
	\caption{Comparison of detection performance between YOLO-Ant and baseline models on the antenna interference source dataset.}
\label{ant_cp}
\end{figure*}

 \vspace{-2pt}
\noindent {\bf{Validity of the proposed module for other models:}}
In our ablation experiment, we demonstrated that modifying the model structure and adding two modules, DSLK-Block and DSLKVit-Block, can effectively improve the baseline model in terms of parameter size, computational complexity, and accuracy. To further demonstrate the effectiveness and applicability of these two modules, we conducted various model comparisons on the antenna dataset. After each model was tested on the antenna dataset, we added the two modules in turn to the model before further comparison; the experimental results are shown in Table \ref{tab:4} and Fig. \ref{zhexian}(b). The two modules we proposed perform well on various commonly used detectors, and the improvement effect they bring is consistent with the experimental trend on the baseline. The introduction of the DSLK-Block to replace the original convolution modules resulted in a significant improvement in the lightweight nature of all the models. However, due to differences in the complexity of their respective neck structure designs, there were substantial variations in the change in model parameter count with the introduction of the DSLKVit-Block. Nevertheless, without exception, the introduction of both modules led to a noticeable enhancement in the model's detection accuracy. Fig. \ref{ant_cp} shows a comparison of the detection performances of our model and the baseline model on the antenna dataset. The DSLK-Block proved instrumental in detecting numerous small targets, particularly patch antennas, which the baseline model failed to identify. When faced with the common background of balcony fences, the baseline model is prone to mistaken local regions for the Yagi antenna, resulting in many false alarms. When encountering the problem of large interclass differences presented by plate log antennas in multiangle shooting, the baseline model’s detection resulted in many false positives. To address these issues, our proposed method combines a CNN and a transformer, introducing the DSLKVit-Block module to effectively resolve the problem. A comparison of the detection figures clearly reveals that our method results in fewer false alarms and misses after the correct targets are detected. Finally, we compare the performance of YOLO-Ant with that of many classical and cutting-edge detectors on the antenna dataset in three dimensions, namely, the model parameter number, accuracy, and computational complexity, as shown in Fig. \ref{zhexian}(c), and YOLO-Ant achieves the best performance.

\begin{figure*}[!t]
	\centering
	\includegraphics[scale=0.82]{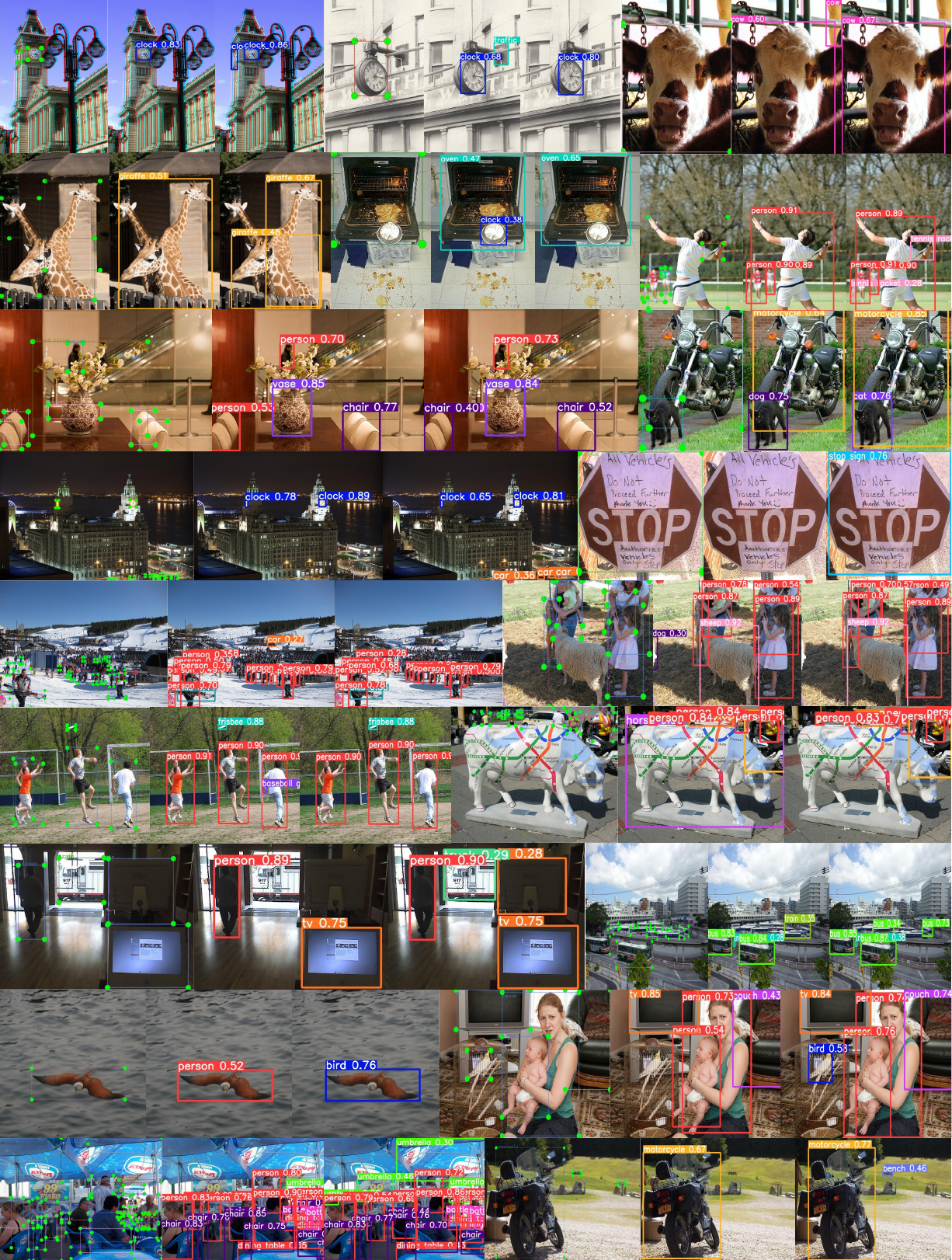}
	\caption{Comparison of detection performance between YOLO Ant and baseline models in the COCO dataset. The three images from left to right in each group represent the ground truth, the detection effect of baseline and YOLO-Ant, respectively.}
\label{coco_cp}
\end{figure*}

\noindent {\bf{Experiments on Public Datasets:}}
To demonstrate the generalizability, robustness, and capability of YOLO-Ant, which can excel not only on antenna datasets but also on public datasets, we conducted comparisons with numerous models on two common datasets, COCO and VisDrone. As shown in Tables \ref{tab:5} and \ref{tab:6}, we compare our model with several other recent outstanding one-stage detectors and several lightweight transformer-based detectors. The results indicate that YOLO-Ant has the fewest parameters and relatively lower computational complexity, outperforming other transformer-based detection models by a significant margin. In terms of detection accuracy, our approach slightly lags behind the top-performing PPYOLO-s in terms of mAP.5 and slightly lags behind PPYOLO-s and DAMO-YOLO-tiny in terms of mAP.5:.95. However, in small object detection, our approach achieves the highest precision among all the models. Fig. \ref{coco_cp} shows the detection performance comparison between our model and YOLOv5-s on the COCO dataset. On the VisDrone dataset, YOLO-Ant also achieves competitive results. Overall, considering the parameter count, computational complexity, and detection accuracy, YOLO-Ant is one of the top contenders among lightweight object detectors.

\noindent {\bf{Model Detection Speed Test:}}
In this experiment, we conducted a speed comparison between YOLO-Ant and several transformer models, as well as YOLO series models known for their lightweight design. We converted all the model weights trained on the antenna dataset into ONNX format and tested them using the OPENVINO tool on a platform with low-computational power(Intel $Core^{TM}$ i9-11900 CPU). Testing was performed on the validation set of the antenna dataset. The comparative results are presented in Table \ref{tab:7}.

\begin{table}[!t]
	\centering
	\caption{Model Detection Speed Test}
	\label{tab:7}  
	\begin{tabular}{ccccc}
		\toprule 
		model & Param.(M) & GFLOPs & FPS\\
		\hline 
		YOLOv5-s \cite{glenn_jocher_2021_5563715}&7.23 & 16.5 & 32.43 \\
            YOLOv7-tiny \cite{wang2022yolov7}&6.23 & 13.86 & 43.82 \\
            DETR(ResNet50)\cite{carion2020end}&41.58&79.52&3.89\\
            RT-DETR(ResNet18)\cite{lv2023detrs} &20.00 & 60.00 & 10.89 \\
            \hline 
		ours & 6.1 & 16.2 & 35.87 \\
		\bottomrule 
	\end{tabular}
 \vspace{-12pt}
\end{table}

The table shows that YOLOv7-tiny achieves an FPS (frames per second) rate exceeding 40, making it the fastest model. However, this lightweight design comes at the cost of reduced detection accuracy. On the other hand, while YOLO-Ant exhibits a slightly slower detection speed than YOLOv7-tiny does, it outperforms the baseline model YOLOv5-s to a small extent. Additionally, the lightweight design of the proposed model is significantly superior to that of the two transformer-based models. This demonstrates the success of our structural design, both in terms of accuracy and lightweight efficiency.

Due to the extensive use of depthwise separable convolutions in our designed modules, which require more memory bandwidth, the I/O (input/output) read speed of the device became the speed bottleneck of the model. We believe that in future work, further optimizations can be made in this regard. This approach enables YOLO-Ant to achieve not only the highest detection accuracy in antenna detection but also further enhance its lightweight design.

\section{Conclusion}
Taking UAV detection of antenna interference sources as the starting point, we propose a lightweight object detector with improved YOLOv5. Initially, to ensure the model is lightweight and adaptable to subsequent modifications, the new network is first pruned based on YOLOv5, which effectively reduces the number of model parameters and computational complexity while ensuring accuracy. To address the challenges posed by small target sizes and complex backgrounds in antenna detection tasks, we propose an efficient and lightweight convolutional module called DSLK-Block. Furthermore, we introduce a lightweight transformer structure integrated with DSLK-Block, which is applied to the network's neck. This combination significantly enhances the network's ability to extract and process features. The new model not only is effective on the antenna dataset but also achieves competitive results on public datasets such as COCO. In future work, we will further explore the integration of our current efforts with drone inspection technology. Simultaneously, we will incorporate traditional equipment such as spectrum analyzers required for conventional antenna interference source identification. The subsequent  objective will be to refine the current approach to create a comprehensive intelligent unmanned detection system.

\bibliographystyle{IEEEtran}  
\bibliography{IEEEabrv,YOLO-Ant}
\newpage
\begin{IEEEbiography}[
	{\includegraphics[width=1in,height=1.25in,clip,keepaspectratio]{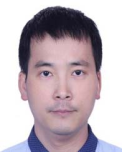}}]{Xiaoyu Tang}
	(Member, IEEE) received the B.S. degree from South China Normal University in 2003 and the M.S. degree from Sun Yat-sen University in 2011. He is currently pursuing the Ph.D. degree with South China Normal University. He is working with the School of Physics, South China Normal University, where he engaged in information system development. His research interests include machine vision, intelligent control, and the Internet of Things. He is a member of the IEEE ICICSP Technical Committee.
\end{IEEEbiography}
\begin{IEEEbiography}[
	{\includegraphics[width=1in,height=1.25in,clip,keepaspectratio]{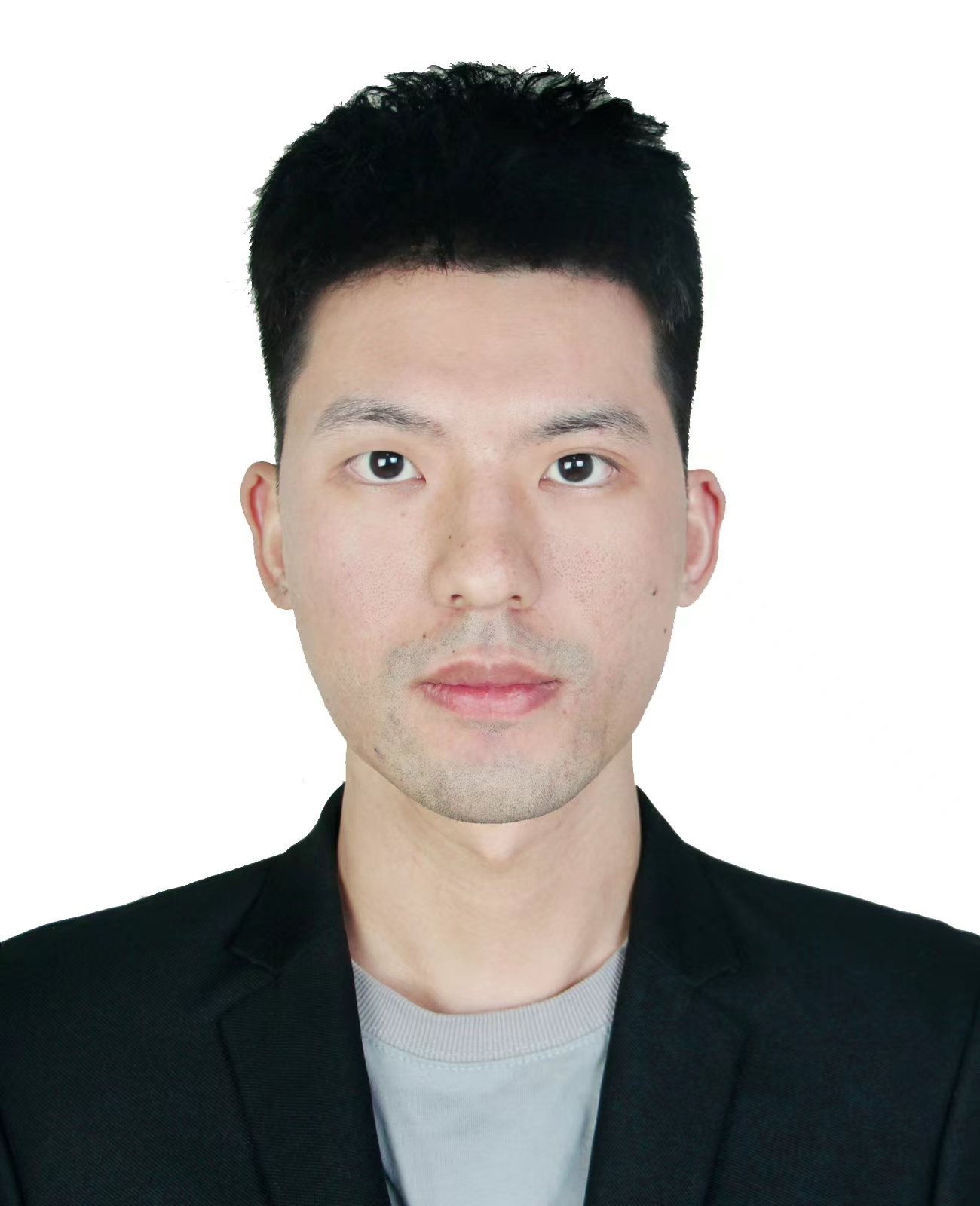}}]{Xingming Chen}
	received his B.Eng. degree from the School of Physics and Telecommunication Engineering at South China Normal University in 2021. Currently, he is pursuing the M.E. degree with the Department of Electronics and Information Engineering at the same institution. His research focuses on computer vision and deep learning.
\end{IEEEbiography}

\begin{IEEEbiography}[
	{\includegraphics[width=1in,height=1.25in,clip,keepaspectratio]{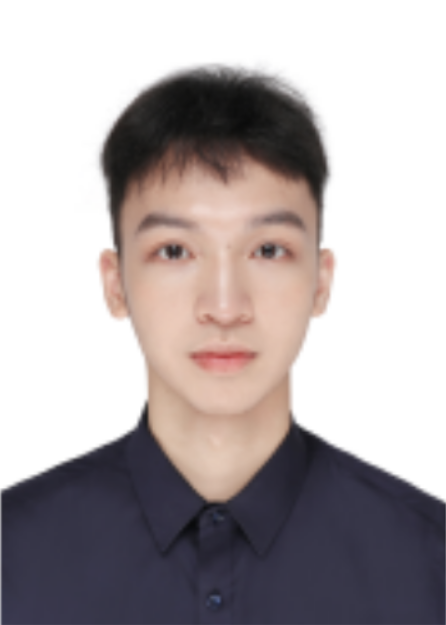}}]{Jintao Cheng}
    received his B.Eng. degree from the School of Physics and Telecommunications Engineering at South China Normal University in 2021. His research focuses on computer vision, SLAM, and deep learning.
\end{IEEEbiography}

\begin{IEEEbiography}[
	{\includegraphics[width=1in,height=1.25in,clip]{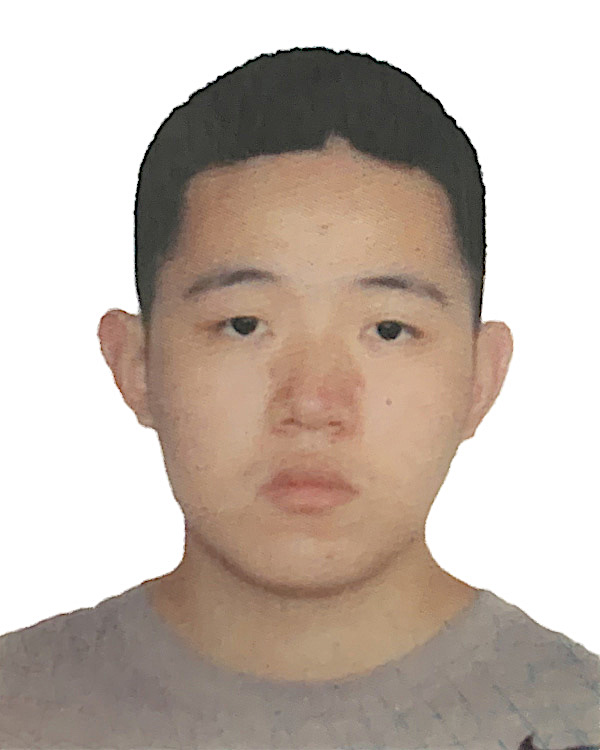}}]{Jin Wu}
(Member, IEEE) was born in Zhenjiang, China, in 1994. He received a B.S. degree from the University of Electronic Science and Technology of China, Chengdu, China. From 2013 to 2014, He was a visiting student with Groep T, Katholieke Universiteit Leuven (KU Leuven). He is currently pursuing a Ph.D. degree in the Department of Electronic and Computer Engineering, Hong Kong University of Science and Technology (HKUST), Hong Kong. He has co-authored over 120 technical papers in representative journals and conference proceedings. He was awarded the outstanding reviewer of \textsc{IEEE Transactions on Instrumentation and Measurement} in 2021. He is now a Review Editor of Frontiers in Aerospace Engineering and an invited guest editor for 5 special issues of MDPI. He is also in the IEEE Consumer Technology Society (CTSoc), as a committee member and publication liaison. He was a committee member for the IEEE CoDIT conference in 2019, a special section chair for the IEEE ICGNC conference in 2021, a special session chair for the 2023 IEEE International Conference on Intelligent Transportation Systems (ITSC), a Track Chair for the 2024 IEEE International Conference on Consumer Electronics (ICCE) and a Chair for the 2024 IEEE CTSoc Gaming, Entertainment and Media (GEM) conference. He was selected as the World’s Top 2$\%$ Scientist by Stanford University and Elsevier, in the 2020, 2021 and 2022 year round.
         \vspace{10pt} 
\end{IEEEbiography}

\begin{IEEEbiography}[{\includegraphics[width=1in,clip]{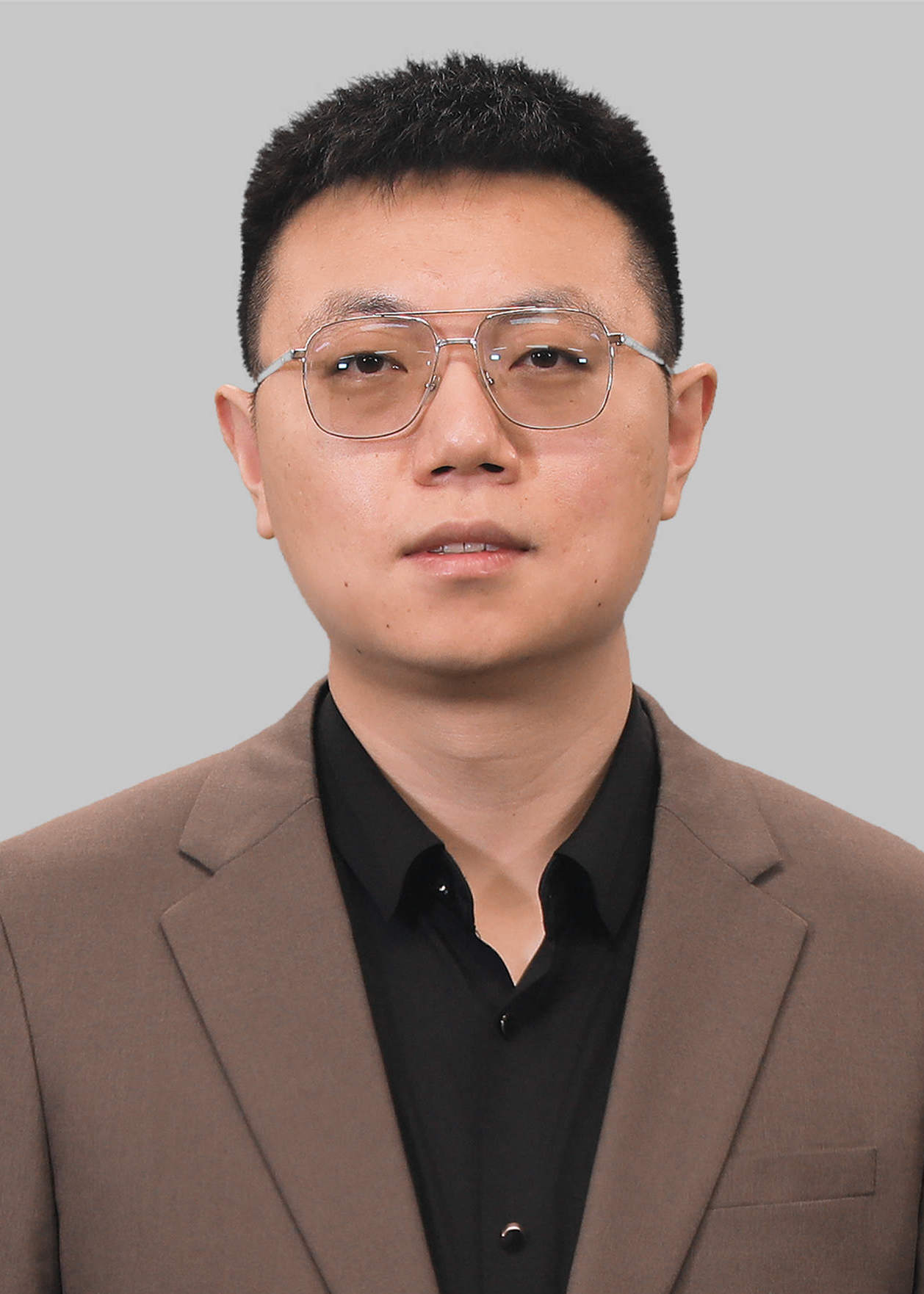}}]{Rui Fan}(Senior Member, IEEE) received the B.Eng. degree in Automation from the Harbin Institute of Technology in 2015 and the Ph.D. degree (supervisors: Prof. John G. Rarity and Prof. Naim Dahnoun) in Electrical and Electronic Engineering from the University of Bristol in 2018. He worked as a Research Associate (supervisor: Prof. Ming Liu) at the Hong Kong University of Science and Technology from 2018 to 2020 and a Postdoctoral Scholar-Employee (supervisors: Prof. Linda M. Zangwill and Prof. David J. Kriegman) at the University of California San Diego between 2020 and 2021. He began his faculty career as a Full Research Professor with the College of Electronics \& Information Engineering at Tongji University in 2021, and was then promoted to a Full Professor in the same college, as well as at the Shanghai Research Institute for Intelligent Autonomous Systems in 2022. 

Prof. Fan served as an Associate Editor for ICRA'23 and IROS'23/24, an Area Chair for ICIP'24, and a Senior Program Committee Member for AAAI'23/24. He is the general chair of the AVVision community and organized several impactful workshops and special sessions in conjunction with WACV'21, ICIP'21/22/23, ICCV'21, and ECCV'22. He was honored by being included in the Stanford University List of Top 2\% Scientists Worldwide in both 2022 and 2023, recognized on the Forbes China List of 100 Outstanding Overseas Returnees in 2023, and acknowledged as one of Xiaomi Young Talents in 2023. His research interests include computer vision, deep learning, and robotics. 
\end{IEEEbiography}

\begin{IEEEbiography}[
	{\includegraphics[width=1in,height=1.25in,clip,keepaspectratio]{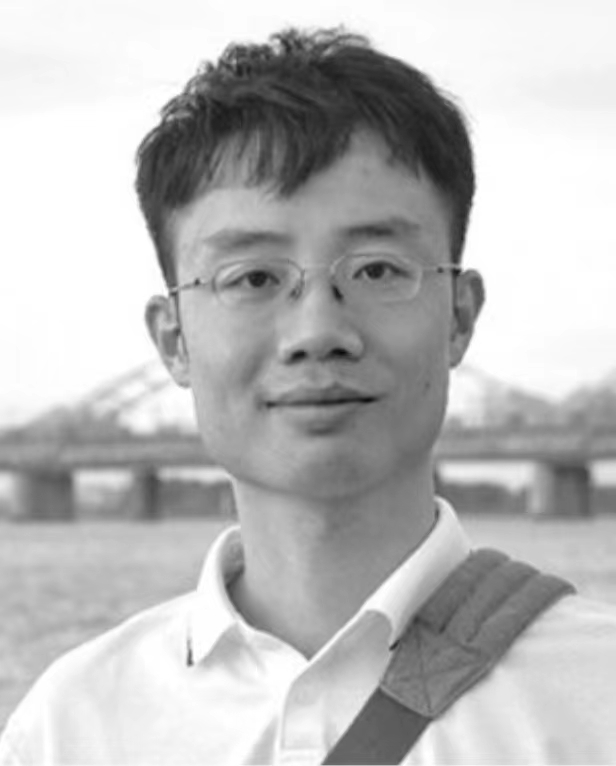}}]{Chengxi Zhang}
	(Member, IEEE) was born in Shandong, China, in February 1990. He received the B.S. and M.S. degrees in microelectronics and solid-state electronics from Harbin Institute of Technology, Harbin, China, in 2012 and 2015, respectively, and the Ph.D. degree in control science and engineering from Shanghai Jiao Tong University, Shanghai, China, in 2019.
	
	He was a Postdoctoral Fellow with the School of Electronic and Information Engineering, Harbin Institute of Technology (Shenzhen Campus),
	Shenzhen, China, from 2020 to 2022. His research interests are embedded system software and hardware design, information fusion, and control theory.
\end{IEEEbiography}

\begin{IEEEbiography}[
	{\includegraphics[width=1in,height=1.25in,clip]{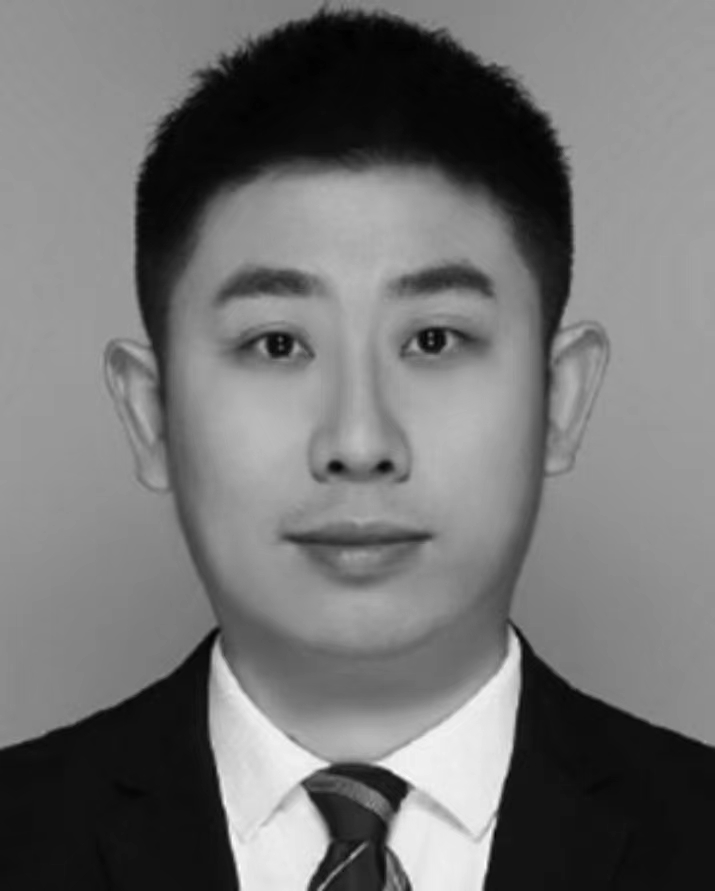}}]{Zebo Zhou}
	was born in November, 1982 in Yongchuan, Chongqing, China. He received the B.Sc. and M.Sc. degrees from the School of Geodesy and Geomatics, Wuhan University, Wuhan, China, in 2004 and 2006, respectively, and the Ph.D. degree from the College of Surveying and Geoinformatics, Tongji University, Shanghai, China, in 2009. In 2009 and 2015, he was a Visiting Fellow with Surveying and Geospatial Engineering Group, within the School of Civil and Environmental Engineering, University of New South Wales, Kensington, NSW, Australia. He is currently an Associate Professor with the School of Aeronautics and Astronautics, University of Electronic Science and Technology of China, Chengdu, China. His research interests include GNSS navigation and positioning, GNSS/INS integrated navigation, and multisensor fusion. 
	
	He has been an In Charge of projects of the National Natural Science Foundation of China and has taken part in the National 863 High-tech Founding of China. He was the Guest Editor of several special issues published on \emph{International Journal of Distributed Sensor Networks and Asian Journal of Control}. He has been presenting related works on the annual conference of the institute of navigation, the annual Chinese satellite navigation conference for several times and has received the best paper awards in these conferences.
\end{IEEEbiography}
\end{document}